\documentclass[12pt]{article}
\usepackage{graphicx}
\usepackage{natbib} 
\usepackage{url} 

\newcommand{\blind}{1}

\newcommand{\zenodoLink}{\url{https://doi.org/10.5281/zenodo.11032650}}

\usepackage{wrapfig}

\def\spacingset#1{\renewcommand{\baselinestretch}%
{#1}\small\normalsize} \spacingset{0}

\usepackage[english]{babel}
\usepackage{amsthm}
\newtheoremstyle{colon}%
{}
{}
{\itshape}
{}
{\bfseries}
{:}
{ }
{}

\theoremstyle{colon}

\newtheorem{theorem}{Theorem}

\newtheorem{prop}{Proposition}

\newcommand{\MD}{{\mathit{SAnD}}}
\newcommand{\MDext}{{Simple Anomaly Detection}}

\usepackage{numprint}
\newcommand{\np}[1]{\numprint{#1}}
\newcommand{\celli}{A.Celli}

\usepackage{scalerel,stackengine}			
\usepackage{amssymb}
\usepackage{authblk}
\usepackage{amsmath, amsfonts, amssymb, bm, bbm}
\usepackage{graphicx}
\usepackage{listings}
\usepackage{float}
\usepackage[titletoc]{appendix}
\usepackage{hvfloat}
\usepackage{longtable}
\usepackage{mathtools}
\usepackage{booktabs,xcolor,colortbl}
\usepackage{xparse}
\usepackage[bottom]{footmisc}
\usepackage[font=small,skip=0pt]{caption}
\usepackage{mathrsfs}
\usepackage{enumitem}
\usepackage{rotating}
\usepackage{comment}
\usepackage{accents}
\usepackage{csquotes}
\usepackage{booktabs}
\usepackage{threeparttable}
\usepackage{setspace}
\excludecomment{confidential}
\let\originalleft\left
\let\originalright\right
\renewcommand{\left}{\mathopen{}\mathclose\bgroup\originalleft}
\renewcommand{\right}{\aftergroup\egroup\originalright}

\newcommand{\tnew}{\tilde{t}}

\usepackage{thmtools}
\declaretheoremstyle[
  bodyfont=\normalfont\itshape,
  headformat=\NAME{\ }A{}\NUMBER
]{nospacetheorem}
\declaretheorem[style=nospacetheorem,name=Assumption{}]{assumption}

\newcommand{\Assone}{$\mathbf{A1}$}

\usepackage{algorithm}
\usepackage{algpseudocode}

{
    \def\OldComma{,}
    \catcode`\,=13
    \def,{%
      \ifmmode%
        \OldComma\discretionary{}{}{}%
      \else%
        \OldComma%
      \fi%
    }%

\begin{document}

\def\spacingset#1{\renewcommand{\baselinestretch}%
{#1}\small\normalsize} \spacingset{1}


\newcommand{\sandTitle}{Accurate and fast anomaly detection in industrial processes and IoT environments}

\if1\blind
{
  \title{\bf \sandTitle}
\author{
	Simone Tonini, Andrea Vandin
	\thanks{	\textit{This work was supported by the Fsc regional Tuscan project AUTOXAI2 J53D21003810008.}}
		\\
    L'EMbeDS \& Inst. of Economics,
    Sant'Anna School for Advanced Studies\\
    \smallskip
     Francesca Chiaromonte \\
       Dept. of Statistics, 
     The Pennsylvania State University \\
     L'EMbeDS \& Inst. of Economics, 
     Sant'Anna School for Advanced Studies  \\
    \smallskip
      Daniele Licari, L'EMbeDS, Sant'Anna School for Advanced Studies
 \\
        \smallskip
         Fernando Barsacchi,    A. Celli Group - Lucca 
    }
\date{}
\maketitle
} \fi

\if0\blind
{
  \bigskip
  \bigskip
  \bigskip
  \begin{center}
    {\LARGE\bf \sandTitle}
\end{center}
  \medskip
} \fi

\begin{abstract}
We present a novel, simple and widely applicable semi-supervised procedure for anomaly detection in industrial and IoT environments, $\MD{}$ (\MDext). $\MD{}$ comprises 5 steps, each leveraging well-known statistical tools, namely; smoothing filters, variance inflation factors, the Mahalanobis distance, threshold selection algorithms and feature importance techniques. To our knowledge, $\MD$ is the first procedure that integrates these tools to identify anomalies and help decipher their putative causes. We show how each step contributes to tackling technical challenges that practitioners face when detecting anomalies in industrial contexts, where signals can be highly multicollinear, have unknown distributions, and intertwine short-lived noise with the long(er)-lived actual anomalies. The development of $\MD{}$ was motivated by a concrete case study from our industrial partner, which we use here to show its effectiveness. We also evaluate the performance of $\MD{}$ by comparing it with a selection of semi-supervised methods on public datasets from the literature on anomaly detection. We conclude that $\MD{}$ is effective, broadly applicable, and outperforms existing approaches in both anomaly detection and runtime.
\end{abstract}

\noindent%

\spacingset{2} 

\section{Introduction}\label{sec:intro}
In the era of Industry 4.0 and the \enquote{Internet of Things} (IoT), industries collect massive amounts of data 
on many aspects 
of their production processes -- usually in the form of time series concerning a large number of variables. This data helps domain experts detect potentially {\it anomalous} behaviors (production errors, technical problems, system defects, breakdowns, outages) which can make 
processes inefficient or sub-optimal. Despite a large literature, anomaly detection remains a particularly complex task (see, e.g., the recent review in~\citealt{ Schmidl2022}). The main difficulty lies in identifying approaches that guarantee at the same time broad applicability, good performance, and low computational cost. Several characteristics of industrial time series can create challenges in anomaly detection. In particular, industrial settings are often characterized by confounding environmental effects (e.g., vibrations, temperature, humidity) which induce a variability of the same magnitude as that due to actual anomalies, complicating identification of the latter~\citep{peeters2001,peeters2001_2,alampalli2000,Deraemaeker2018,Grosskopf2022}. 
This limits the range of anomaly detection methods that can be successfully implemented in industrial settings.

Our contribution consists of a novel semi-supervised anomaly detection procedure, named $\MD{}$ (\MDext), which leverages a combination of well-known statistical tools (smoothing filters, variance inflation factors, the Mahalanobis distance, threshold selection algorithms, and feature importance techniques). The development of our procedure was motivated by a concrete case study offered by an industrial partner. We use such case study to show the effectiveness 
of our procedure in practice. Furthermore, we evaluate performance by comparing our procedure with the semi-supervised methods reviewed in~\cite{Schmidl2022}. The comparison is conducted on 8 public datasets from the anomaly detection literature, representing a broad range of potential applications. This demonstrates the generality of our approach. Overall, we conclude that our procedure is effective, broadly applicable, and outperforms existing approaches in both anomaly detection and runtime. The fact that a pipeline based on relatively simple tools can outperform more sophisticated methods is, in fact, in line with considerations in the review by ~\cite{Schmidl2022}, who note that \emph{simple} methods tend to yield performance comparable to that of more elaborate counterparts, at lower computational costs.

\begin{figure}[t]
	\graphicspath{{images/}}
	\centering
	\subfloat[]{\includegraphics[width=5cm]{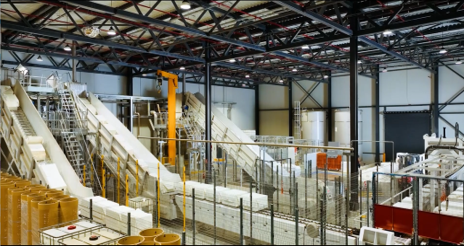}}\hfil
	\subfloat[]{\includegraphics[width=5cm]{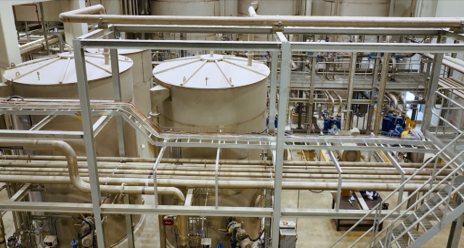}}\hfil 
	\subfloat[]{\includegraphics[width=5cm]{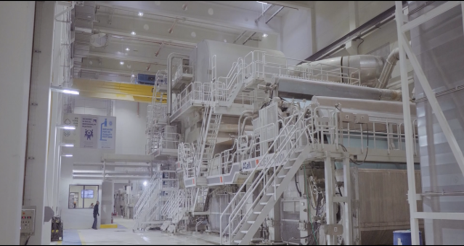}}\hfill
	
	\subfloat[]{\includegraphics[width=6.28cm]{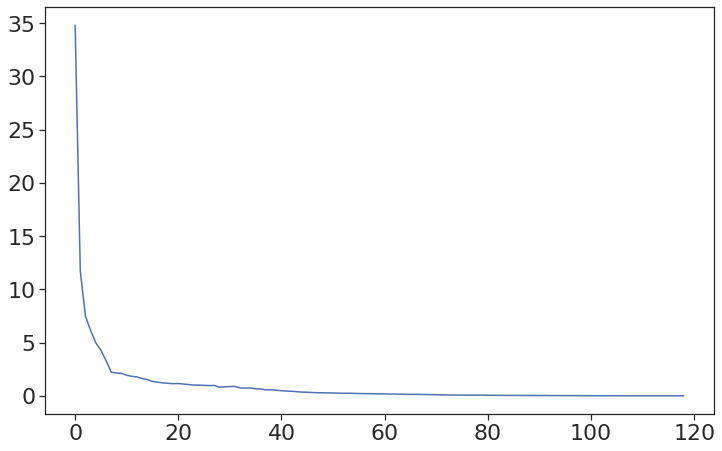}}\hfil 
	\subfloat[]{\includegraphics[width=6.4cm]{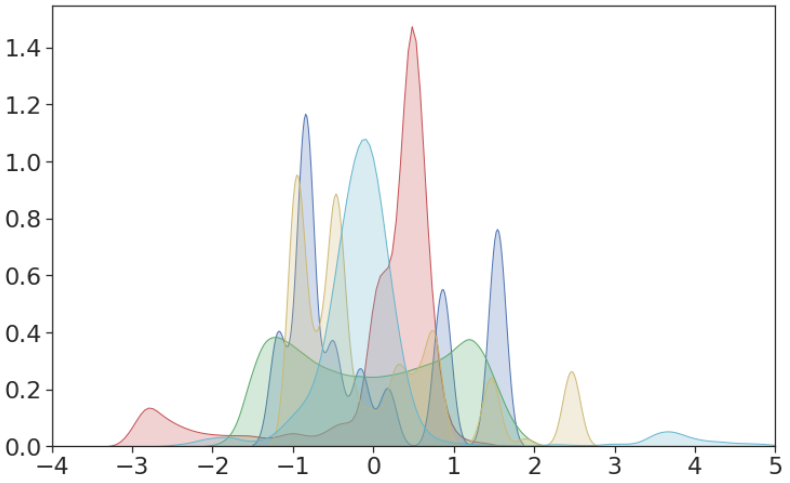}}\hfil
	\caption{Panels
		(a)-(c) show three phases of the production process of the \celli{} tissue machine. 
		Panel (d) shows the eigenvalues of the correlation matrix of the 119 variables considered in our case study. 
		Panel (e) shows the densities of 5 among these variables after standardization.  
	}\label{fig:CelliChallenges}
\end{figure}

$\MD{}$  was developed with the industrial partner \celli{}, leader in the supply of machinery and technologies for the paper and nonwoven products industry. The company provided data  from a tissue paper machine, which comprised 217 variables measured every second for about 19 hours. This is a complex production line with several phases, three of which are depicted in Figure~\ref{fig:CelliChallenges} (a)-(c), that can 
tackle different 
products. We focused on a specific product, which reduced the 
variables to consider to 119.
The collaboration aimed to develop a methodology to detect anomalies in the production process of
tissue machines, with two key requirements, namely: ({\bf R1}) enable quick and accurate identification of the time intervals in which
anomalies occur; ({\bf R2}) provide information on patterns and variables involved in the anomalies, to help domain experts decipher their 
causes. 

We note that the data at our disposal contained exclusively intervals assessed as \emph{anomaly-free} by the domain experts. In other words, no known instances of anomalies were provided, so we could not resort to supervised methods, which need to be trained on data comprising both anomalies and anomaly-free observations. The procedure we developed classifies as \emph{semi-supervised} based on the definition in~\cite{Schmidl2022}; we do not exploit \emph{anomaly} labels in the training data, but rather data concerning only \emph{normal}, anomaly-free behavior.

In addition to {\bf R1} and {\bf R2}, the considered scenario 
required that we focus on {\em long-lived anomalies}, i.e., anomalies lasting for minutes (over observations taken per second), since in this domain short-lived anomalies are likely due to sensor noise.
Note that, for each long-lived anomaly, it suffices to flag just one observation  within its time interval to alert an expert who will then investigate the issue. In other words, long-lived anomalies can be detected through at least one true positive within their time-span. 
Finally, 
we wanted our procedure to be  general, and thus easily exportable to additional 
(industrial) domains.

The last decade has seen the development of a multitude of sophisticated semi-supervised methods based, e.g., on deep learning or statistical learning techniques, which should be able to tackle {\bf R1-2} by leveraging complex patterns in the data without the introduction of strong assumptions (see~\citealt{choi2021, blazquez2021review, Schmidl2022} for recent reviews). However, the comprehensive comparison in~\cite{Schmidl2022} demonstrates that no single method offers the best performance across different scenarios. 

%
In contrast, $\MD{}$ exploits the {\em Mahalanobis distance}~\citep{mahalanobis1936} evaluated 
on the training data as a means to identify anomalies. This is an extremely simple approach, which requires only estimates of location and covariation 
from the training data. However, data from industrial processes do pose technical challenges ({\bf TC}) also for the use of such a straightforward approach. The first ({\bf TC1}) is multicollinearity. Figure \ref{fig:CelliChallenges} (d) shows the eigenvalues of the correlation matrix of the 119 variables generated by the \celli{} tissue machine; the smallest are close to 0, indicating strong multicollinearity. This is a problem, since the Mahalanobis distance requires the covariance 
matrix to be invertible. 
The second technical challenge ({\bf TC2}) concerns the distribution of the data. Figure \ref{fig:CelliChallenges} (e) shows the densities of 5 variables randomly selected among the 119. Clearly, even under anomaly-free conditions, the distributions are far from Gaussian, or other typical forms, e.g., used to represent skewed data~\citep{Peter1990,Hubert2008,Moti2010,TODESCHINI2013,Cabana2019}. 
When one can postulate that anomaly-free data follow a Gaussian or other known distributions, it is possible to derive a distribution for the Mahalanobis distance and, based on such distribution, to define a threshold for the indentification of anomalies~\citep{rousseeuw2005robust,maronna2019robust}. When distributional assumptions are unsuitable, one must resort to different approaches to define an appropriate threshold.

\begin{figure}[t]
	\graphicspath{{images/}}
	\centering
	{\includegraphics[width=0.8\linewidth]{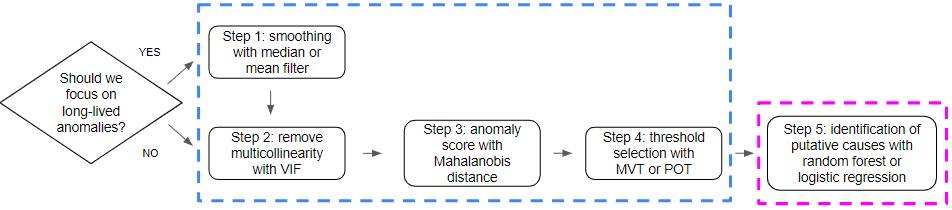}}
	\caption{\footnotesize Flow chart of the proposed 
		procedure. The dashed blue line shows the steps used to
  tackle {\bf R1}, while the dashed purple line shows the step used to
  tackle {\bf R2}.}\label{buildingblocks2}
\end{figure}

Our procedure, which attempts to address all the challenges and considerations raised thus far, consists of the 5 steps depicted in Figure \ref{buildingblocks2}. We start with an anomaly-free dataset, which represents the training set for our procedure. The first step consists of smoothing the data through a median or mean filter, as to remove short-lived anomalies; this step is optional, and should be applied only if the focus is on long-lived anomalies. The second step mitigates multicollinearity removing variables based on their {\em variance inflation factors}~\citep{VIF2002}. In the third step, we compute the {\em Mahalanobis distances} for all points in the training set. The fourth step provides a {\em threshold} for the identification of anomalies. If no distributional assumptions are possible, we use the maximum value of the Mahalanobis distances in the training set (MVT), or the value produced by a Peaks-Over-Threshold (POT) analysis of the right tail of such distances~\citep{balkema1974,pickands1975}. The data in the test set undergoes smoothing, as the training data. A Mahlanobis distance is produced for each test data point, and a (binary) anomaly prediction is produced based on whether its distance is above threshold. 
Finally, the fifth step uses feature importance techniques offered by random forest~\citep{Breimanrf01} or logistic regression (\citealt{James2013}, ch. 4) to analyze anomalies predicted in the test set, unveiling patterns and pinpointing relevant variables, as to decipher their putative causes.

In our comparisons, we employ common performance metrics such as the Matthews correlation coefficient (MCC)~\citep{chicco2020}, F1, recall, and precision. We also assess computational cost through runtime analyses. Additionally, we showcase the effectiveness of our procedure on the \celli{} case study which motivated its development.

The article is organized as follows. 
Section~\ref{theory} focussz on the properties of the Mahalanobis distance. 
Section~\ref{Met} details $\MD$. 
Section~\ref{sec:labelledDATAexp} demonstrates its performance through a comprehensive validation on methods and datasets from~\cite{Schmidl2022}. Section~\ref{sec:rootCause}  
relates its feature importance results
to the causes of the anomalies.
Finally, Section~\ref{sec:val_case_study} demonstrates $\MD{}$ on the case study, and Section~\ref{sec:conc} provides  final remarks. 
Replicability material for Sections~\ref{sec:labelledDATAexp} and~\ref{sec:rootCause} is available at \zenodoLink{}.

\section{Some background on the Mahalanobis distance}
\label{theory}
The Mahalanobis distance~\citep{mahalanobis1936} is one of the key
statistical 
tools employed by $\MD$.
It has low computational requirements and high interpretability, as it allows one to quantify how much an observation deviates from ``behavior under normal conditions'' through straightforward 
estimation of the
mean vector and covariance matrix of the training data.
Here we focus on aspects 
that make it suitable for anomaly detection in semi-supervised contexts, while we discuss other background material 
in 
Supplement \ref{othertheory}.

Let $\mathbf{X}_A = \left\{\mathbf{x}_{t}\right\}_{t=1}^{T_A}$ denote an $n \times T_A$ rectangular array of observations on $n$ stationary time series (i.e., time series with finite variance, see~\citealt{choi2021}) with $T_A$ observations for each series.
In practice, $\mathbf{X}_A$ 
is provided by a domain expert to denote the behavior of the process under normal conditions.  
Given a new rectangular array $\mathbf{X}_B = \left\{\mathbf{x}_{t}\right\}_{t=T_{A}+1}^{T}$ of dimension $n\times T_B$, where $T_B=T-T_A$, we focus on the question: 
\emph{How distant is each point in $\mathbf{X}_{B}$ from the 
behavior captured by $\mathbf{X}_A$?}
The Mahalanobis distance is a common approach to answer this. 
Let $\widehat{\pmb{\mu}}_A$
and 
$\widehat{\pmb{\Sigma}}_A$ 
be the 
mean vector and covariance matrix estimated on $\mathbf{X}_A$. 
For a given $n$-dimensional observation in the test set, say $\mathbf{x}_{\tnew}=(x_{1,\tnew},\dots,x_{n,\tnew})'$ 
with $T_{A}\!+\!1\leq\tnew\leq T$,
the Mahalanobis distance 
from the training set is defined as
\begin{equation}\label{MahaDist}
MD_{\tnew}=\sqrt{(\mathbf{x}_{\tnew}-\widehat{\pmb{\mu}}_A)'\ \widehat{\pmb{\Sigma}}_A^{-1}\ (\mathbf{x}_{\tnew}-\widehat{\pmb{\mu}}_A)} \ \ .
\end{equation}
$MD_{\tnew}$ 
measures how far $\mathbf{x}_{\tnew}$ is from the center of the training data in an inner product that is shaped by its
inverse covariance -- so directions of higher (co)variation matter less in assessing departure from the center. 
%
%

As mentioned in the Introduction, 
industrial processes can have
large operational variability, as well as substantial collinearities, under normal conditions. The next assumption is fundamental to apply semi-supervised anomaly detection methods in this context.

\begin{assumption}\label{ass:1}
$\mathbf{X}_A$ is representative of the operational variability of the process under normal conditions.    
\end{assumption}

We can expect \Assone{} to hold for large $T_A$. The importance of \Assone{}  
lies in the fact that we can identify $p\leq n$ eigenvectors of $\widehat{\pmb{\Sigma}}_A$ whose span approximates the linear sub-space to which the normal conditions belong. In particular, let $\mathbf{v}_1,\dots,\mathbf{v}_n$ be the eigenvectors of $\widehat{\pmb{\Sigma}}_A$
ordered based on the corresponding eigenvalues $\lambda_1 \geq \dots
\geq \lambda_n$,
and let $p$ 
be the smallest integer such that $\frac{\sum_{i=1}^p\lambda_i}{\sum_{i=1}^n\lambda_i}>\alpha$, where $\alpha \in [0,1]$
is a threshold value (e.g., 0.99).
This means that 
the first $p$ principal components
explain $100 \times \alpha$\% of the variability of $\mathbf{X}_A$. 
Under \Assone, 
such components explain most of the operational variability of the process under normal conditions, while the Mahalanobis distance will increase 
more steeply for an observation varying in the direction of the remaining $n-p$ components~\citep{Deraemaeker2018}. The following proposition formalizes this reasoning.

\begin{prop}\label{prop:1}
Under Assumption \Assone{}, let $\sum_{i=1}^p\lambda_i$ be the sum of the $p$ largest eigenvalues,  associated with the eigenvectors that explain $100 \times \alpha\%$ of the variability of $\mathbf{X}_A$. 
If $\sum_{i=1}^p \lambda_i\rightarrow\sum_{i=1}^n \lambda_i$ (i.e., $\alpha\rightarrow\ 1^-$), then $MD_{\tnew}$ tends to the Mahalanobis distance of $\mathbf{x}_{\tnew}$ projected on the space described by the remaining $n-p$ components.
\end{prop}

\noindent\textbf{Proof:} See Supplement \ref{prop1:proof}. 

Proposition \ref{prop:1} shows that by including all operational variability (the variables measured in all possible environmental conditions) in computing the covariance matrix $\widehat{\mathbf{\Sigma}}_A$, the Mahalanobis distance increases as $\mathbf{x}_{\tnew}$ moves away from the normal conditions of the process. 
%
This property justifies its use in anomaly detection. 
However, as discussed in 
the Introduction, 
some technical challenges (\textbf{TC1-2}) 
complicate such use
in the considered domain. Section \ref{Met} shows how some of the
step included in $\MD{}$ solve these issues.

\section{The 5-step \emph{SAnD} procedure
}
\label{Met}
In this section, we present our procedure to detect anomalies in industrial contexts. $\MD{}$ 
combines five main elements; 
namely:
smoothing filters, variance inflation factors, Mahalanobis distance, threshold selection procedures, and feature importance techniques. Let $\mathbf{X}_A$ and $\mathbf{X}_B$ be rectangular arrays  on $n$ variables as in Section~\ref{theory}.  
For the $n\times T$ matrix $\mathbf{X}=(\mathbf{X}_A,\mathbf{X}_B)$, the proposed procedure 
comprises the following 
steps.

\paragraph{Step 1 - Smoothing (optional).} If 
the problem at hand concerns only long-lived anomalies,
we smooth each of the time series
with a filter based on location measures 
(mean, or median for robustness) computed on a moving window of given size, say
$h$. 
This replaces each of the $T$-dimensional vectors $\mathbf{x}_i$, $i=1,\dots,n$ 
with a corresponding $(T-(h-1))$-vector of smoothed values
$\mathbf{w}_i$. 
To simplify notation, 
we indicate with $\ddot{T}=T-2(h-1)$, $\ddot{T}_A=T_A-(h-1)$
and $\ddot{T}_B=T_B-(h-1)$ the sizes of the smoothed vectors. Therefore, 
we replace the $n\times T$ matrix $\mathbf{X} = (\mathbf{X}_A,\mathbf{X}_B)$ with the $n\times\ddot{T}$ matrix $\mathbf{W}=(\mathbf{W}_A,\mathbf{W}_B)$, where $\mathbf{W}_A$ 
($n\times\ddot{T}_A$) 
comprises smoothed 
values in normal conditions, and $\mathbf{W}_B$ 
($n\times\ddot{T}_B$) 
smoothed values 
among which we want to detect anomalies.  

%

Smoothing allows us to 
mitigate short-lived anomalies that would be
considered as sensor noise by domain experts, and thus to avoid false positives.  
Domain experts will be in a position to select between mean filtering or the more robust median one, and to specify an appropriate window size $h$ for given applications.
In the experiments  in the following Sections, we 
rely on median filtering and we consider $h=1$ (no smoothing) and $h=10$. 

\paragraph{Step 2 - Removing multicollinearity (to address {\bf TC1}).} 
We use 
Variance Inflation Factors, VIF~\citep{VIF2002}, to quantify 
multicollinearity among the variables in the training data $\mathbf{W}_A$ (or $\mathbf{X}_A$ if we omit the smoothing) and iteratively remove some of them. 
We first subtract from the entries of each $\mathbf{w}_{Ai}$ its mean $\bar{w}_{Ai}=\ddot{T}_A^{-1}\sum_{t=1}^{\ddot{T}_A}w_{it}$, switching to $\mathbf{d}_{Ai}=\mathbf{w}_{Ai}-\bar{w}_{Ai} \mathbf{1}$, and form the centered matrix $\mathbf{D}_A$ 
(centering is not necessary here; however, popular statistical software packages require it 
in order to compute VIFs 
ignoring
intercepts). Next, we regress each variable in $\mathbf{D}_A$ against all others, compute the coefficients of determination from such regressions, say $R_i^2$, 
and thus the VIFs, which are given by
VIF$_i\!=\!1/(1-R_i^2)$, $i\!=\!1,\ldots n$.
We remove the variable with largest VIF, re-run the regressions and recompute the VIFs, remove again the variable with  largest VIF, {\em et cetera} -- 
until the largest VIF is less than 5, a 
benchmark commonly used in the literature~(\citealt{James2013}, ch. 3.3). 
This leaves us with a set of $m$ $(\leq n)$ variables with at most mild multicollinearity. We 
indicate the reduced, centered training data matrix as $\widetilde{\mathbf{D}}_A$.

%
Mitigating multicollinearity is critical because the Mahalanobis distance utilizes 
the inverse of the covariance matrix estimated on the data 
(see Equation \eqref{MahaDist}); the strong linear associations which exist in our case study 
(see Figure \ref{fig:CelliChallenges} (d)) and many other applications may prevent such inversion.
The covariance matrix $\pmb{\tilde{\Sigma}}_A=\ddot{T}_A^{-1}\widetilde{\mathbf{D}}_A\widetilde{\mathbf{D}}_A'$ relative to the $m$ non-multicollinear variables will not present any invertibility issues.

\paragraph{Step 3 - Computing 
anomaly scores.} 
We start by processing the test data $\mathbf{W}_B$. First, we center it with the means computed on the training data, switching to $\mathbf{d}_{Bi}=\mathbf{w}_{Bi}-\bar{w}_{Ai} \mathbf{1}$ and forming the centered matrix $\mathbf{D}_B$. Next, we reduce such matrix eliminating the same variables that were eliminated from the training data in Step 2; we indicate the reduced, centered test data matrix as $\widetilde{\mathbf{D}}_B$. 
Finally, we calculate 
Mahalanobis 
distances for 
the full dataset $\widetilde{\mathbf{D}}=(\widetilde{\mathbf{D}}_A,\widetilde{\mathbf{D}}_B)$ (that is, for both training and test observations) using the training covariance matrix; in symbols 
$MD_t=\sqrt{\widetilde{\mathbf{d}}_t'\pmb{\tilde{\Sigma}}_A^{-1}\widetilde{\mathbf{d}}_t}$,
$t=1,\dots,\ddot{T}$. We indicate with  
$\pmb{MD}_T$, 
$\pmb{MD}_A$ and $\pmb{MD}_B$, respectively, the $T$,
$\ddot{T}_A$
and $\ddot{T}_B$-dimensional vectors of anomaly scores for the full dataset, the training set and the test set.

\paragraph{Step 4 - Threshold selection (to address {\bf TC2}).} 
We use $\pmb{MD}_A$ to select a threshold  $k$ beyond which an observation in $\pmb{MD}_B$ is 
flagged as an anomaly. Note that if we can assume the 
variables retained in Step 2 to be distributed as an $m$-variate Gaussian under normal conditions, the quadratic form expressing the Mahalanobis distance will be distributed as a chi-square with $m$ degrees of freedom~\citep{penny1996}. Thus, setting the threshold as a quantile of such chi-square, say $k=\chi^2_m(1-\alpha)$, would guarantee a p-value of $\alpha$ when flagging anomalies (one-sided rejections).
%
However, unfortunately, the variables 
in our reference domain are far from Gaussian
(see Figure \ref{fig:CelliChallenges} (e)). 
We therefore resort to other approaches to select the threshold $k$; namely, the {\em Maximum Value in the Training sample} (MVT) and {\em Peaks-Over-Threshold} (POT) approaches. 
In the
former, the threshold is simply set at the largest value in $\pmb{MD}_A$,	using, in a way, an empirical p-value of 0 when flagging anomalies. 
%
In the latter, we fit a generalized Pareto distribution~\citep{Siffer2017} to the \emph{peaks} in the training set, i.e., the elements of $\pmb{MD}_A$ above a given  percentile  (we fix the 99th one). 
Using a known formula (see Supplement \ref{EVTandPOT}), we thus obtain a threshold $k$ that depends on the peaks, on 
the parameters of the fitted distribution, and on a chosen
probability (we fix 0.001) for a peak to be an anomaly,
or an \emph{extreme event} in POT terminology.
%
%
%
%

Once the threshold $k$ is fixed, we compare each observation $MD_{\tnew}$ in $\pmb{MD}_B$ with $k$, and 
generate a binary vector $\widehat{\mathbf{y}}=(\widehat{y}_1,\dots,\widehat{y}_{\ddot{T}_B})$, with $\widehat{y}_{\tilde{t}}= 1$ if $MD_{\tnew}>k$ and 0 otherwise. 
%

\paragraph{Step 5 - 
Identifying variables involved in anomalies (to address {\bf R2}).} After detecting 
anomalies, we try to associate them to
specific variables -- as to aid
domain experts in the investigation of putative causes.
We do so by training supervised methods equipped with feature importance 
techniques, using the detected anomalies in $\widehat{\mathbf{y}}$ as prediction targets. In more detail,
we fit a prediction model on
each time interval of interest (i.e., each interval from the test set corresponding to a detected long-lived anomaly), which we combine with 
arbitrarily selected anomaly-free observations 
(in our experiments in Section~\ref{sec:rootCause} we
use observations from the final portion of the training set, while  in Section~\ref{sec:val_case_study} we 
use anomaly-free observations from the same interval containing anomalies). 
%
We use two well-known 
prediction models;
namely, {\em random forest}~\citep{Breimanrf01} and {\em logistic regression} (\citealt{James2013}, ch. 4). 
Feature importance is evaluated through 
the {\em Gini Index} for the former (see Supplement \ref{RandomForest} and~\citealt{James2013} ch. 8), and through 
the {\em Relative Contribution to Deviance Explained} (RCDE) for the latter
(see Supplement \ref{RCDE} and~\citealt{tripodi2020}).

\section{Evaluation 
of performance and runtime}\label{sec:labelledDATAexp}
We evaluate our procedure in terms of performance and computational cost by comparing it with several
state-of-the-art semi-supervised anomaly detection techniques on multiple public datasets from the anomaly detection literature. For the selection of 
techniques and 
datasets to employ in our comparison, we rely on the recent, broad survey by~\cite{Schmidl2022}.
%
The survey 
comprised eleven 
semi-supervised techniques, of which we consider nine; namely,
LSTM-AD~\citep{malhotra2015}, HealthESN~\citep{chen2020}, Telemanom~\citep{hundman2018}, Random Black Forest~\citep{Schmidl2022}, EncDec-AD~\citep{Malhotra2016}, DEEPAnT~\citep{munir2018}, Omnianomaly \citep{Su2019}, Robust-PCA~\citep{paffenroth2018}, and Hybrid-KNN~\citep{song2017} -- all run using settings and implementations provided in 
\cite{Schmidl2022} and in 
its replicability material at \url{https://github.com/TimeEval/TimeEval-algorithms}.  
We omit LaserDBN~\citep{ogbechie2017} and TAnoGan \citep{bashar2020} due to
implementation issues, but these
were
among the bottom ranking in the evaluation by~\cite{Schmidl2022}.
%
%

We run $\MD$ with both $h=1$ (no smoothing, $\MD_{1}$) and $h=10$ (smoothing, $\MD_{10}$).  
%
Since most of the 
procedures, including ours, provide an \emph{anomaly score} to be compared against a 
threshold, we homogenize the comparison
using the same threshold selection methods, namely MVT and POT, for all procedures. 
The only exception is Hybrid-KNN, which returns the probability that an observation is an anomaly. Here we present results obtained setting a threshold of $0.8$ on such probability; 
higher thresholds (e.g., $0.9$ and $0.99$) do not noticeably change the performance of Hybrid-KNN 
(see Supplement \ref{KNN_others}).

The survey 
by~\cite{Schmidl2022} considered 24 collections of public datasets from the anomaly detection literature. 
Only 2 of these collections contain non-synthetic multivariate time series  and comprise anomaly-free data as required by 
semi-supervised methods; namely, the {\em Server Machine Datasets}, SMD~\citep{Su2019}, and the {\em Exathlon}~\citep{Pavel2016}. 
From each of these two collections we randomly selected 4 datasets that contain long-lived anomalies. 
 SMD, one of the largest public 
data repositories available for anomaly detection in multivariate time-series, contains 5-week-long datasets collected per minute by a large internet company on 28 different machines. It 
comprises one dataset per machine,  
each with 38 variables.  Exathlon contains 39 datasets with about 23 variables on average. It has been used in the context of industrial fault detection, a specific type of anomaly.  For both collections, the \emph{training sets} are anomaly-free, while the test sets are not, and have labels to denote anomalies.  The datasets from SMD contain both short- and long-lived anomalies, while the ones from Exathlon contain only long-lived anomalies.
Table \ref{Tab:DataDesc} shows some statistics on the  datasets.

\begin{table}[t]
\centering
\scalebox{0.75}{
\begin{tabular}{cccc}
\toprule
\emph{Collection} & \emph{Avg. Size} & \emph{Avg. Dim.} & \emph{Avg. Cont.} \\
\midrule
     SMD & \np{54827} & 38.00 & 4.04\%\\
Exathlon & \np{94186} & 23.25 & 7.32\%\\
\bottomrule
\end{tabular}
}             
\caption{\footnotesize 
Some statistics 
on the 8 
datasets used in our comparison. We consider 4 datasets from each of two collections, listing averages for:   
number of observations (\emph{Avg. Size}), number of variables (\emph{Avg. Dim.}), and contamination (\emph{Avg. Cont.}) -- i.e., the
percentage of test observations labeled as anomalies. 
}
\label{Tab:DataDesc}
\end{table}

In terms of performance metrics, we use various well-known indicators of accuracy in anomaly detection; namely:
precision, 
$\mathit{Prec}=TP/(TP+FP)$;
recall, 
$\mathit{Recall}=TP/(TP+FN)$;
the harmonic mean of the two, or {\it F1} score, 
$F1=2TP/(2TP+FP+FN)$;
and the Matthews Correlation Coefficient~\citep{chicco2020}
$$MCC=\frac{TP\times TN-FP\times FN}{\sqrt{(TP+FP)\times (TP+FN)\times (TN+FP)\times (TN+FN)}}$$
which, ranging in $[-1,1]$, measures the 
agreement between true and detected anomalies. All these metrics combine, in intuitive and effective ways, counts of 
true positives (TP), true negatives (TN), false positives (FP) and false negatives (FN).
 %
 %
%
%
Notably though, since they consider individual anomalous observations,
they might not be suited  to evaluate methods when the focus is on long-lived anomalies. 
This is because they do not tell us whether long-lived anomalies were
flagged by detecting at least one anomaly within them. 
To capture this, we use an additional metric named
\emph{RIC (Ratio of Identified Clusters)}, where a ``cluster'', i.e., a long-lived anomaly, is counted as identified
if at least one observation within its time frame is
detected as an anomaly (see Supplement~\ref{RIC}). 

\begin{table}[t]
\centering
\scalebox{0.655}{
\begin{tabular}{crcrccc}
\toprule
\multicolumn{1}{c}{\emph{Method}}  &  \multicolumn{1}{c}{\emph{Prec}} &  \multicolumn{1}{c}{\emph{Recall}} &  \multicolumn{1}{c}{\emph{F1}} &  \multicolumn{1}{c}{\emph{MCC}}   &  \multicolumn{1}{c}{\emph{RIC}}   &  \multicolumn{1}{c}{\emph{\%No Anom.}} \\ 
\midrule
\midrule
\multicolumn{7}{c}{\emph{MVT}}\\
\midrule

   LSTM-AD        &0.667   (0.37)     &0.124   (0.20)    &0.104   (0.15)     & 0.168   (0.15)    &0.309   (0.44)     &62.5\%     \\
  HealthESN       &0.582   (0.39)     &{\bf0.511}(0.51)  &0.211   (0.16)     & 0.209   (0.13)    &{\bf0.928}(0.14)   &\bf0.0\%  \\
  Telemanom       &0.713   (0.17)     &0.054   (0.08)    &0.104   (0.14)     & 0.148   (0.21)    &0.229   (0.37)     &62.5\%     \\
   RBForest       &0.998   (0.01)     &0.014   (0.02)    &0.030   (0.04)     & 0.140   (0.07)    &0.365   (0.45)     &50.0\%     \\
  EncDec-AD       &0.370   (0.52)     &0.003   (0.01)    &0.009   (0.02)     &-0.007   (0.10)    &0.025   (0.07)     &50.0\%     \\
    DeepAnT       &0.827   (0.23)     &0.100   (0.16)    &0.140   (0.22)     & 0.278   (0.22)    &0.246   (0.27)     &62.5\%     \\
 OmniAnomaly      &0.907   (0.09)     &0.093   (0.14)    &0.152   (0.22)     & 0.318   (0.26)    &0.272   (0.38)     &62.5\%     \\
  RobustPCA       &{\bf1.000}(0.00)   &0.033   (0.05)    &0.061   (0.08)     & 0.222   (0.11)    &0.192   (0.28)     &50.0\%   \\
     $\MD_1$      &{\bf1.000}(0.00)   & 0.304  (0.43)    &0.342   (0.44)     & 0.427   (0.43)    &0.813   (0.37)     &12.5\%   \\
     $\MD_{10}$   &0.872   (0.31)     & 0.491  (0.43)    &{\bf0.650}(0.41)   &{\bf0.630}(0.45)   &0.823   (0.35)     &12.5\%   \\
\midrule
\multicolumn{7}{c}{\emph{POT}}\\
\midrule   
    LSTM-AD      &0.603     (0.37)   &0.155    (0.25)   &0.117    (0.14)   &0.160    (0.15)   &0.376    (0.47)    &62.5\%     \\
  HealthESN      &0.529     (0.37)   &0.496    (0.47)   &0.244    (0.12)   &0.219    (0.15)   &{\bf1.000}(0.00)   &\bf0.0\%     \\
  Telemanom      &0.852     (0.17)   &0.460    (0.41)   &0.522    (0.38)   &0.482    (0.41)   &0.638    (0.43)    &16.67\%     \\
   RBForest      &0.614     (0.28)   &0.533    (0.38)   &0.421    (0.28)   &0.369    (0.31)   &{\bf1.000}(0.00)   &\bf0.0\%     \\
  EncDec-AD      &0.573     (0.52)   &0.018    (0.02)   &0.038    (0.04)   &0.035    (0.14)   &0.112    (0.13)    &25.0\%     \\
    DeepAnT      &0.715     (0.32)   &0.266    (0.32)   &0.140    (0.34)   &0.380    (0.37)   &0.469    (0.51)    &50.0\%     \\
OmniAnomaly      &0.883     (0.12)   &0.164    (0.17)   &0.315    (0.23)   &0.320    (0.26)   &0.436    (0.41)    &40.0\%     \\
  RobustPCA      &0.890     (0.09)   &0.061    (0.08)   &0.110    (0.15)   &0.250    (0.14)   &0.475    (0.51)    &50.0\%     \\
     $\MD_1$     &{\bf0.901}(0.17)   &0.585    (0.44)   &0.635    (0.39)   &0.624    (0.41)   &{\bf1.000}(0.00)   &\bf0.0\%   \\ 
    $\MD_{10}$   &0.841     (0.25)   &{\bf0.673}(0.42)  &{\bf0.666}(0.38)  &{\bf0.645}(0.40)  &0.844    (0.35)    &12.5\%   \\

\midrule
\multicolumn{7}{c}{\emph{$\Pr>0.8$}}\\
\midrule
 Hybrid KNN   &0.143 (0.31)  &0.312 (0.43) &0.309 (0.22) &0.959 (0.22)  & 0.625 (0.21) &0.0\%   \\
\bottomrule
\end{tabular}
}     
\caption{\footnotesize Performance evaluation for each method and 
threshold selection procedure. In the POT case, HealthESN and RBForest were run on 7 datasets, Telemanom on 6, and Omnianomaly on 5 because the parameters of the generalized Pareto distribution could not be calculated on some of the datasets (see Supplement~\ref{EVTandPOT}). For the Exathlon collection, EncDec-AD failed to calculate anomaly scores.
In addition to the performance metrics, which are averaged over the datasets (in parenthesis the standard deviations), 
the last column provides the percentage of datasets out of the considered ones where no long-lived anomalies were identified (\emph{\%No Anom.}). 
The best performance for each metric and thresholding procedure is in bold.  
}\label{tab:TabResu}
\end{table}

Table \ref{tab:TabResu} reports, for each method and threshold selection procedure, all
performance metrics averaged over the datasets,
along with the percentage of datasets  
where no long-lived anomalies were identified ({\it \%No Anom.}). 
$\MD$ shows consistently high performance.
In particular, it always has the best 
performance with POT, 
and either the best or second-best with MVT.
Smoothing in $\MD$ tends to have a positive impact on \emph{F1} and \emph{MCC}, while {\it \%No Anom.} might worsen due to the loss of anomalies in the datasets from SMD. 

\begin{figure}[t]
\graphicspath{{images/}}
\centering
{\includegraphics[width=0.65\linewidth]{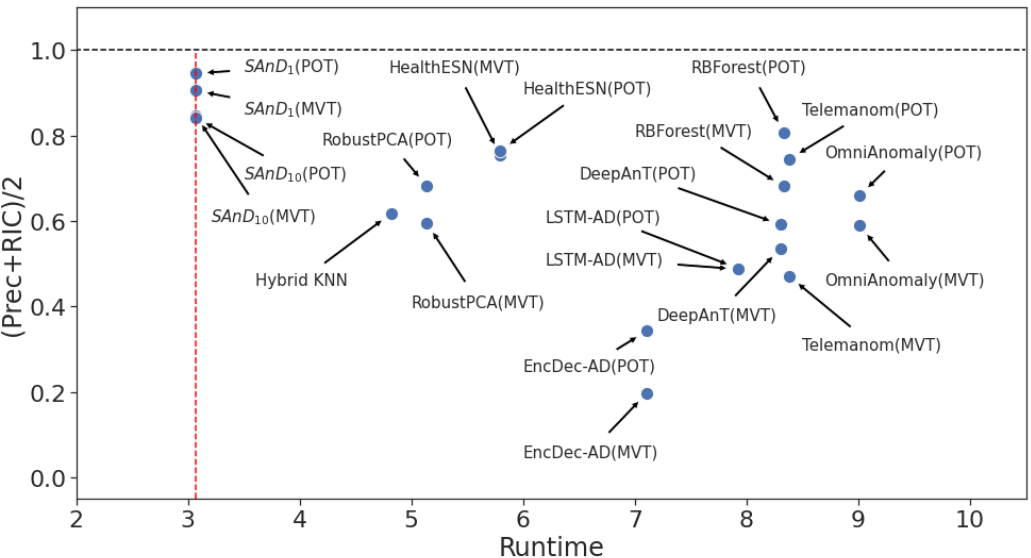}}
\caption{\footnotesize Performance of each method and threshold procedure, focusing on 
long-lived anomalies (y-axis), vs runtime expressed as logarithms of seconds needed to compute anomaly scores (x-axis). Performances and runtimes are averaged across the considered datasets.
Hybrid KNN uses the threshold $\Pr>0.8$.
}\label{fig:RunTimPrecRIC}
\end{figure}

We now focus on 
performance in detecting long-lived anomalies. For each dataset, we compute the arithmetic mean between  
\emph{Prec} 
(which controls false positives) and \emph{RIC} (which targets specifically long-lived anomalies), 
restricting attention to the portion of the test set that 
contains long-lived anomalies (see Supplement~\ref{TestPLOT} for more details).
%
In Figure~\ref{fig:RunTimPrecRIC}, for each method and thresholding procedure combination, the average of this summary across the considered datasets is plotted against
the average runtime of the method -- computed as the logarithm of the seconds needed to compute the anomaly scores.
In general, we see that POT 
improves the identification of long-lived anomalies (see also $RIC$ results 
in Table~\ref{tab:TabResu}; 
the improvement is most marked for the Telemanom method).
Most notably though, with both thresholding procedures and with and without smoothing, 
$\MD$ is located 
in the upper left corner of Figure~\ref{fig:RunTimPrecRIC}, representing the
best-performing method 
with the least runtime. In Supplement \ref{PrecRIC_TravsTest} we report 
runtimes separately for training and test sets. 
%
In summary, our comparison demonstrates that $\MD$
outperforms competitors in both anomaly detection accuracy and runtime, addressing \textbf{R1}.

\section{Evaluation of
feature importance 
}
\label{sec:rootCause}
The last step of $\MD$ uses random forest or logistic regression to associate detected anomalies to specific variables. Here, we try and evaluate 
whether and how feature importance assessed by these prediction models 
can shed light on the putative causes of 
anomalies. 
%
The datasets in SMD 
comprise information on the variables that caused each anomaly. 
We thus use 
the first dataset from SMD for our evaluation, focusing in particular on 5 anomalies, among the 8 within it, that are long-lived
(they last several minutes, see Supplement~\ref{TestPLOT}).  

The first four steps of $\MD$ run with $h=1$ and POT detect all these long-lived anomalies. 
We run step 5 
by training  
random forest and logistic regression for each such anomaly,
using \np{1000} observations from the test set containing the 
anomaly, and \np{1000} additional anomaly-free observations from the training set 
(see Supplement~\ref{TestPLOT} for details).

\begin{table}[t]
	\centering
	\scalebox{0.63}{
		\begin{tabular}{c l l l}
			\toprule
			\multicolumn{1}{c}{\emph{Anom.}} &  \multicolumn{1}{c}{\emph{Causes}} & \multicolumn{1}{c}{\emph{Ranked by RF, $h=1$}} & \multicolumn{1}{c}{\emph{Ranked by LR, $h=1$}}\\
			\midrule
			1  & 1,9,10,12,13,14,15 &{\bf10},{\bf12},{\bf13},{\bf9},{\bf15},{\bf14},{\bf1} & {\bf10},{\bf15},{\bf9},\textcolor{red}{6},{\bf13},\textcolor{red}{2},{\bf1} \\
			2&1,2,3,4,6,7,9,10,11,12, & {\bf34},{\bf20},{\bf35},{\bf1},{\bf36},{\bf13},{\bf10},{\bf28},{\bf29},{\bf31}, & {\bf10},{\bf1},{\bf6},{\bf4},{\bf9},{\bf15},\textcolor{red}{23},{\bf2},{\bf11},{\bf36},\\
			
			& 13,14,15,16,19,20,21,22,24,25,& {\bf12},{\bf22},{\bf2},{\bf21},{\bf3},{\bf33},{\bf25},{\bf19},{\bf15},{\bf14}, & {\bf35},{\bf3},{\bf32},{\bf16},{\bf21},{\bf20},{\bf19},{\bf13},{\bf34},{\bf14},\\
			& 26,27,28,29,30,31,32,33,34,35,36 &{\bf6},{\bf26},{\bf4},{\bf32},\textcolor{red}{23},{\bf16},{\bf7},{\bf11},{\bf24},{\bf30} &{\bf24},{\bf31},{\bf26},{\bf29},{\bf33},{\bf25},{\bf22},{\bf30},{\bf28},{\bf12},{\bf7}\\
			3& 1,2,9,10,12,13,14,15 &{\bf10},{\bf12},{\bf13},\textcolor{red}{20},{\bf1},{\bf9},\textcolor{red}{28},\textcolor{red}{22} & {\bf10},{\bf15},{\bf1},{\bf13},\textcolor{red}{34},\textcolor{red}{29},{\bf9},{\bf14}\\
			4 &1,2,3,4,9,10,11,12,13,14,15,16,25,28 &{\bf10},{\bf12},{\bf13},\textcolor{red}{7},\textcolor{red}{6},\textcolor{red}{20},\textcolor{red}{34},{\bf9},\textcolor{red}{29},\textcolor{red}{36},{\bf25},{\bf28},\textcolor{red}{30},{\bf15} &  {\bf10},\textcolor{red}{6},\textcolor{red}{29},{\bf14},{\bf11},\textcolor{red}{23},{\bf16},{\bf13},\textcolor{red}{34},\textcolor{red}{33},{\bf2},{\bf9},{\bf1},{\bf15}\\
			5  &1,9,10,12,13,14,15&{\bf10},{\bf13},{\bf12},{\bf9},{\bf15},{\bf14},\textcolor{red}{23} & {\bf10},{\bf9},\textcolor{red}{33},{\bf15},\textcolor{red}{23},\textcolor{red}{24},\textcolor{red}{26} \\
			\bottomrule
		\end{tabular}
	}
	\caption{\footnotesize 
		Variables labeled as causes in the dataset vs variables ranked as important for the anomalies by Random Forest (\emph{RF}) and logistic regression (\emph{LR}). We mark such important variables 
  in bold if they are among the labeled causes, and in red if they are not. With very few exceptions (e.g., anomaly 4, LR) the top portion of the rankings is occupied by variables labeled as causes. 
  }\label{Tab:VarSel}
\end{table}

Table~\ref{Tab:VarSel} 
summarizes 
results.
%
The \emph{real} causes
indicated by the dataset description are reported in the section labeled \textit{Causes}; 
variables here are arbitrarily numbered based on the column order in the data set, and listed based on such numbers -- but SMD does not provide
information on the importance of the causes; that is, 
the variables are not ranked.
Sections
\emph{Ranked by RF} and \emph{Ranked by LR} report the
variables identified as most 
relevant using random forest 
and logistic regression, respectively, 
ranked by (decreasing) importance 
(see Supplement~\ref{FeatureImportance}). 
For each anomaly $a$, we 
include the 
$v(a)$ most important variables, 
$v(a)$ being the number of variables indicated 
as causes in the dataset description. 
We 
see that the results of step 5
are indeed coherent with the
causes as given in the dataset; 
82.1\%  and 79.1\% of the variables identified by our procedure using \emph{RF} and \emph{LR}, respectively, are among
such causes. 
This demonstrates how
$\MD$ 
can effectively address \textbf{R2}.

\section{Case study
}\label{sec:val_case_study}
Next, we demonstrate the  effectiveness of $\MD{}$ on an
industrial case study
involving 
a tissue machine of the \celli{} company. This application was 
performed in close 
collaboration with domain experts
who validated 
its results.

The data
consisted of \np{70000} observations collected per second on 217 variables.
We used the first \np{60000} for training, and the last \np{10000} for testing. In the considered time interval, the machine 
handled only one type of product, allowing us to restrict attention to 119 variables deemed relevant for this one product.
%
For the smoothing in step 1 of $\MD{}$, we considered two 
window sizes; $h=1, 10$.
%
For the feature importance assessment in step 5, we focused on random forest
($RF$ performed slightly better than $LR$ in Section~\ref{sec:rootCause}).

\begin{figure}[t]
	\graphicspath{{images/}}
	\centering
	\subfloat[]{\includegraphics[height=2.2cm]{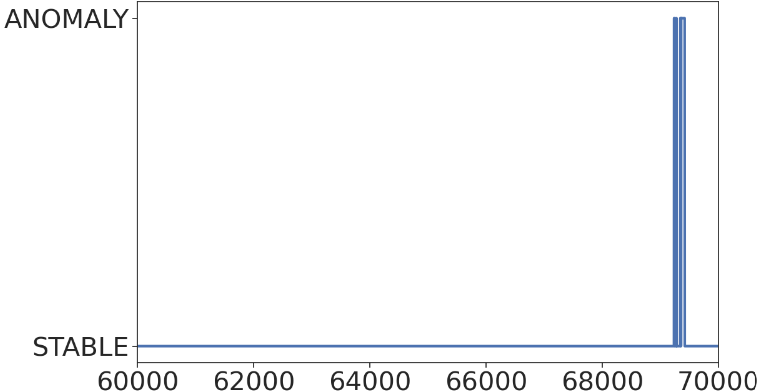}}
	\subfloat[]{\includegraphics[height=2.2cm]{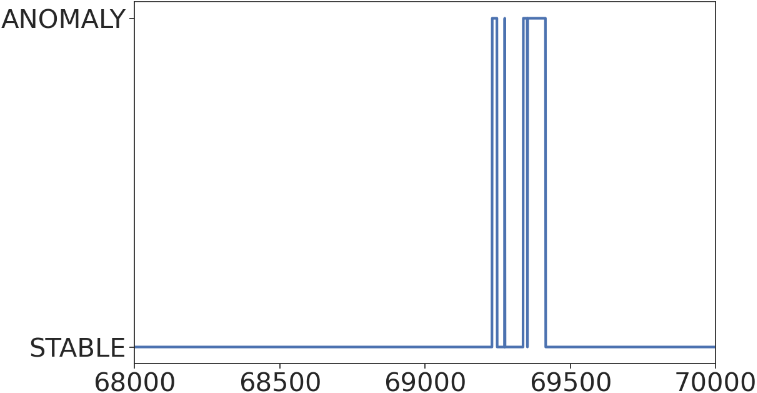}}
	\subfloat[]{\includegraphics[height=2.2cm]{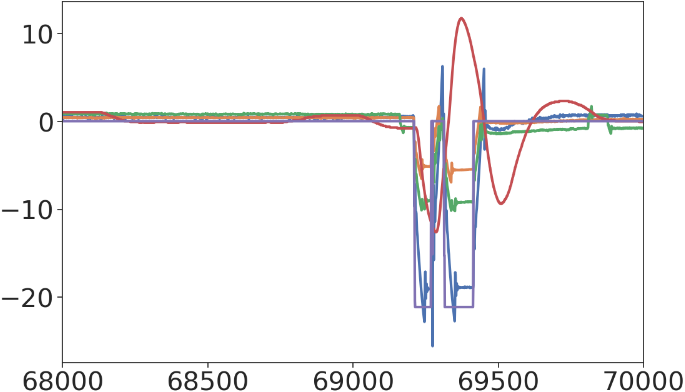}}  
	\subfloat[]{\includegraphics[height=2.2cm]{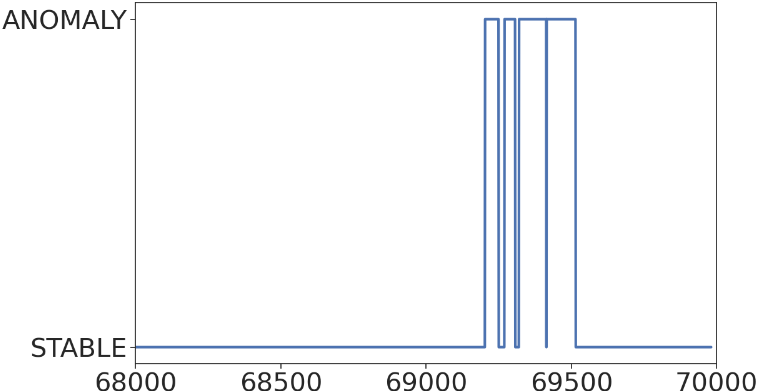}} 
	\caption{\footnotesize Case study
		results using MVT thresholding. Plot (a) shows 
		anomalies detected in the test set by $\MD_1$ (no smoothing, $h=1$). Plot (b) highlights 
		such anomalies zooming in
		on the last \np{2000} observations.
		Plot (c) shows the time series of the 5 most important variables 
		as identified by a 
		Random Forest. Plot (d) is the same as Plot (b), but for $\MD_{10}$ (smoothing with $h=10$).
	}\label{fig:Celli1}
\end{figure}

We start from the analysis with no smoothing 
($h=1$). In step 2, because of high multicollinearity in our case study, 
VIFs reduced the variables from 119 to 60.
In step 3 we computed the anomaly scores, and in step 4 we thresholded them with MVT to detect the four  anomalies shown in Figure \ref{fig:Celli1} (a)-(b).
%
%
Zooming in (Figure \ref{fig:Celli1} (b)), we can see that 
three  are short-lived 
and one is long-lived. 
However, given that they are very close, the domain experts  interpreted the `cluster' of four detected anomalies 
as a single long-lived anomaly.
Therefore, in step 5 we 
investigated the putative causes of the cluster of detected anomalies as a whole. We did this by training 
a random forest on the entire interval 
(2000 observations) shown in Figure \ref{fig:Celli1} (b).
Differently from  Section~\ref{sec:rootCause}, here we do not need to add anomaly-free observations from the training sample, because 
such interval contains plenty of 
anomaly-free 
observations.
%
%
%
The domain experts deemed the top 
$v=5$ variables as sufficient and relevant for 
interpreting the causes of the anomalies. These are: \textit{WERecirculationFanPower}, \textit{DERecirculationFanPower} (two
energy expenditure variables 
for the fans that manage 
humidity and the drying process of the sheet), \textit{MCC1PowerConsumption} (power consumption of the electrical distribution panel), \textit{DEGasConsumption} (gas consumption 
of the drying process),
\textit{V6-Speed} (speed of a  fan used to dry the sheet). 
Figure \ref{fig:Celli1} (c), which depicts 
the time series of 
these 5  variables,
clearly suggests the presence of
a long-lived anomaly.
Indeed, 
the domain experts were able to validate our findings as
an actual process anomaly
caused by overheating of the system during the paper drying phase. 

Running $\MD$ with smoothing ($h=10$) lends further support to the existence of one long-lived anomaly. 
The VIFs in step 2 leads to the same 60 variables obtained without smoothing. 
Thresholding the resulting anomaly scores, again with MVT, detects
one, essentially uninterrupted long-lived anomaly in the same time interval
(Figure~\ref{fig:Celli1} (d)).
A random forest, trained again on the \np{2000} observations in Figure \ref{fig:Celli1} (b)
but using smoothed data,
ranks as top 5 variables 
\textit{BRConsistency} (pulp consistency setpoint),
\textit{WERecirculationFanPower},
\textit{MCC1PowerConsumption},
\textit{Thermocompressor} (sensor on drying phase),
and \textit{SF-Refiner-Outlet-Pressure} (outlet pressure of the SF refiner). 
While three of these variables differ from those 
for $h=1$, according to the domain experts they point to the same cause, i.e., overheating during the paper drying phase.

\begin{figure}[t]
 	\graphicspath{{images/}}
	\centering
	\subfloat[]{\includegraphics[height=2.2cm]{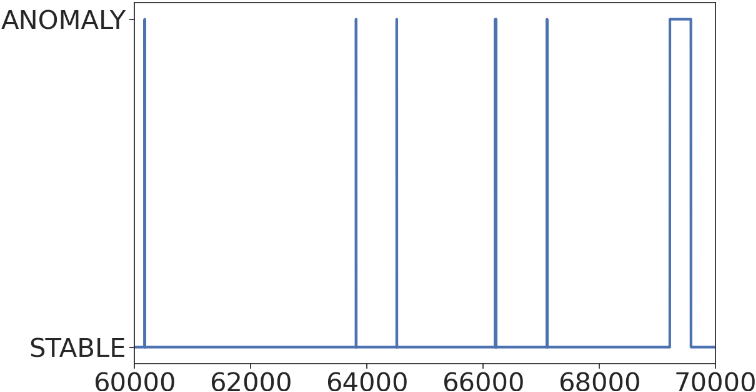}} 
	\subfloat[]{\includegraphics[height=2.2cm]{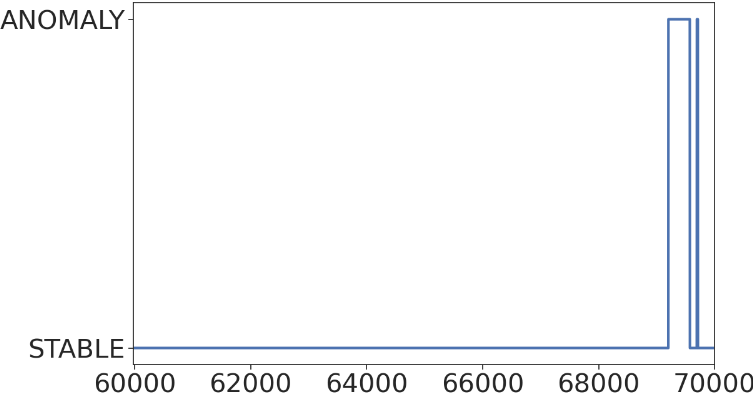}}
 	\subfloat[]{\includegraphics[height=2.2cm]{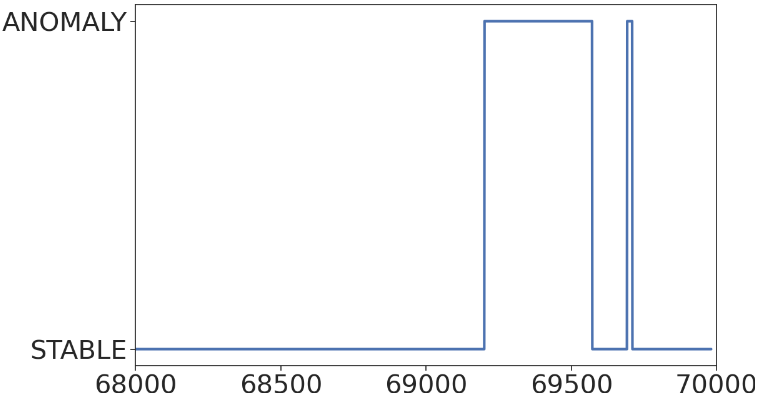}}
	\subfloat[]{\includegraphics[height=2.2cm]{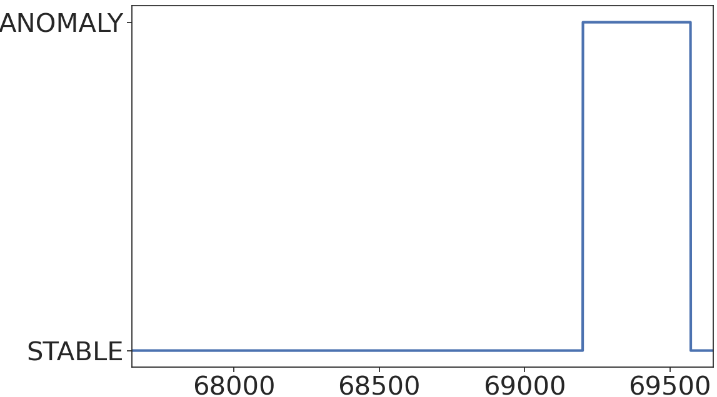}}
	\caption{\footnotesize Results in the case of POT. Plots (a) and (b) show the anomalies detected in the test set by $\MD_{1}$ and $\MD_{10}$, respectively. Plot (c) highlights the anomalies detected by $\MD_{10}$ by showing the last \np{2000} observations only. Plot (d)  highlights the long-lived anomaly detected by $\MD_{10}$.
 }\label{fig:Celli2}
\end{figure}

Next, we switched from MVT to POT thresholding. Figure \ref{fig:Celli2} (a) displays results without smoothing ($h=1$). POT detects many more anomalies than MVT (Figure~\ref{fig:Celli1} (a)), and especially many short-lived ones attributed by domain experts to sensor noise. Figures \ref{fig:Celli2} (b)-(c) display results with smoothing ($h=10$), which makes $\MD$ more suited to long-term anomalies;
$\MD_{10}$ detects 1 long-lived and 1 short-lived anomaly, discarding other sensor noise 
detected by $\MD_{1}$. The domain experts deemed the long-lived anomaly to be the same as that in Figure \ref{fig:Celli1} (d). Indeed, the time interval involved is almost identical.
Training a random forest on the
\np{2000} observations in Figure \ref{fig:Celli2} (d) 
we find that 3 of the 5 top ranked variables are shared with those identified by $\MD_{10}$ 
with MVT thresholding; namely, 
\textit{BRConsistency},
\textit{WERecirculationFanPower} and \textit{MCC1PowerConsumption}. 
The other top variables are 
\textit{SheetON} (if paper passes from the last roll to the winder), and
\textit{DERecirculationFanPower} (also 
identified by $\MD_{1}$ with MVT).
%
According to the domain experts, also this new selection of variables 
pointed to system overheating. 
Step 5
run for the short-lived anomaly in Figure~\ref{fig:Celli2} (c)
produces a different ranking of the variables, but still a clear connection to system overheating (see Supplement~\ref{ShortPOTh10}).

This analysis exemplifies an important aspect of anomaly detection: the \emph{real} cause 
underlying an anomaly might not be directly expressed by the observed variables, and 
domain knowledge is required to reconstruct causes. 
In our case study, 
the variable \enquote{system overheating} does not exist as such. However, the variables selected by step 5 of $\MD$ with various specifications (with or without smoothing, with different thresholding) 
all point to the slowing and cooling of the production system, allowing our industrial partners to identify system overheating as the underlying cause.
%




\section{Concluding remarks}\label{sec:conc}
We 
introduced $\MD$, a procedure based on simple and well-known statistical tools to detect anomalies in industrial processes 
characterized by multicollinear time series with unknown distribution. 
Experiments 
show that $\MD$ outperforms state-of-the-art methods both in terms of 
performance 
and  runtime. $\MD$ also allows us to 
identify relevant variables, shedding light on putative causes underlying anomalies.
Thus, $\MD$ meets 
key requirements for anomaly detection in
industrial settings; namely, to enable domain experts to quickly and accurately identify time intervals when
anomalies occur, and to help decipher their causes.
%
$\MD$ also meets the
flexibility, reliability
and simplicity requirements
recently highlighted, e.g., in
\cite{Schmidl2022}.
In 
our experiments, $\MD$ 
outperformed competing methods in a broad variety of datasets and domains (flexibility)
and it successfully discovered anomalies in all such datasets (reliability). Moreover, employing simple tools, 
$\MD$ does not require convoluted procedures for parameter tuning (simplicity). 

We envision a number of 
avenues for future work. 
One among them is extending $\MD{}$ with the ability to rank anomalies by their impact on domain-specific key performance indicators (KPIs).
%
%
Another is further improving the effectiveness of
step 5 with techniques to counteract the effects of unbalanced 
counts of 
anomalous and anomaly-free observations in the training of prediction models (random forests or logistic regression)~\citep{fithian2014}.
The results of step 5 of the case study illustrated in
Section~\ref{sec:val_case_study}, where the time intervals used to train random forests contained more anomaly-free than anomalous observations, were validated by domain experts. Therefore, $\MD$ appeared not to be 
affected by the unbalanced training, 
but this might not hold in general.

\bibliographystyle{chicago}

\bibliography{Bibliography-MM-MC}

\bigskip
\begin{center}
{\large\bf SUPPLEMENTARY MATERIAL}
\end{center}
\appendix
\normalsize
\begin{appendices}

\section{Proof}\label{prop1:proof}
The spectral decompositions of $\widehat{\pmb{\Sigma}}_A$ and $\widehat{\pmb{\Sigma}}^{-1}_A$ are $\mathbf{V}\pmb{\Lambda}\mathbf{V}'$ and $\mathbf{V}\pmb{\Lambda}^{-1}\mathbf{V}'$, respectively. We now focus on the following transformation of the observed variables: $\pmb{\xi}_t=\mathbf{V}'\mathbf{x}_t$. The corresponding estimated vector of means and covariance matrix are given by
 \begin{align*}
   \widehat{\pmb{\mu}}_{\xi} & =\frac{1}{T}\sum_{t=1}^T\pmb{\xi}_t = \mathbf{V}'\widehat{\pmb{\mu}}_A,\\
   \widehat{\pmb{\Sigma}}_{\xi} & =\cfrac{1}{T-1}\sum_{t=1}^T (\pmb{\xi}_t-\widehat{\pmb{\mu}}_{\xi}) (\pmb{\xi}_t-\widehat{\pmb{\mu}}_{\xi})'=\\
   & = \mathbf{V}'\frac{1}{T-1}\sum_{t=1}^T (\mathbf{x}_t-\widehat{\pmb{\mu}}_A) (\mathbf{x}_t-\widehat{\pmb{\mu}}_A)'\mathbf{V} = \mathbf{V}'\widehat{\pmb{\Sigma}}_A\mathbf{V} = \pmb{\Lambda}.
\end{align*}
The last equality follows by the orthogonality properties and shows that the standard deviation of each component ${\xi}_i$ is given by $\sqrt{\lambda_i}$. Using the inverse transformation $\mathbf{x}_t=\mathbf{V}\pmb{\xi}_t$, we can see that the square of the Mahalanobis distance ($MD_t^2$) reduces to
\begin{equation}\label{MD_eigen}
    MD_t^2=\sum_{i=1}^p\frac{1}{\lambda_i}(\xi_{ti}-\mu_{\xi i})^2.
\end{equation}
Equation \ref{MD_eigen} shows that $MD_t^2$ can be decomposed into a sum of independent contributions from each component of the transformed variables $\xi_{ti}=\mathbf{V}_i'\mathbf{x}_t$. It is noteworthy that the contributions are weighted by the inverse of the associated eigenvalues $\lambda_i$, which are the variances of the new transformed variables. This implies that if the variance is large, the contribution to $MD_t^2$ is small. Thus, for a new observation vector $\mathbf{x}_{\tilde{t}}$, and the corresponding transformation $\pmb{\xi}_{\tilde{t}}=\mathbf{V}'\mathbf{x}_{\tilde{t}}$, we can decompose the square of the Mahalanobis distance into two parts,
\begin{equation}
    MD_{\tilde{t}}^2=\sum_{i=1}^p\frac{1}{\lambda_i}(\xi_{\tilde{t}i}-\mu_{\xi i})^2+\sum_{i=p+1}^n\frac{1}{\lambda_i}(\xi_{\tilde{t}i}-\mu_{\xi i})^2=MD_{1,\tilde{t}}^2+MD_{2,\tilde{t}}^2,
\end{equation}
where $MD_{1,\tilde{t}}^2$ is the square Mahalanobis distance of $\mathbf{x}_{\tilde{t}}$ projected on the principal components, while $MD_{2,\tilde{t}}^2$ is the square Mahalanobis distance of $\mathbf{x}_{\tilde{t}}$ projected on the remaining $n-p$ components. Let $\left(\sum_{i=1}^p\lambda_i/\sum_{i=1}^n\lambda_i\right)>\alpha.$ Thus, we have that for $\sum_{i=1}^p\lambda_i\rightarrow\sum_{i=1}^n\lambda_i$, (i.e., for $\alpha\rightarrow1^-$), $\sum_{i=p+1}^n\lambda_i\rightarrow0$ and therefore $MD_{\tilde{t}}^2\rightarrow\sum_{i=p+1}^n\frac{1}{\lambda_i}(\xi_{\tilde{t}i}-\mu_{\xi i})^2=MD_{2,\tilde{t}}^2.$ The proof ends by considering that $\sqrt{MD_{\tilde{t}}^2}=MD_{\tilde{t}}\rightarrow\sqrt{MD_{2,\tilde{t}}^2}=MD_{2,\tilde{t}}$.

 \hfill $\blacksquare$


\section{Other Statistical Tools}\label{othertheory}

\subsection{Threshold identification with EVT and POT}\label{EVTandPOT}
Extreme value theory (EVT) is a family of techniques that seeks to assess, from a given ordered sample of a given random variable, the probability of events that are more extreme than any previously observed. 
EVT is based on the theoretical result of~\cite{fisher1928} and~\cite{gnedenko1943}, which states that, under a weak condition, extreme events have the following kind of distribution, regardless of the original one,
\begin{equation}\label{FirstEVT}
G_{\gamma}\rightarrow x \text{ exp } \left(-(1+\gamma x)^{-\frac{1}{\gamma}}\right),
\end{equation}
where $\gamma\in\mathbb{R}$ and $1+\gamma x>0$. All the extreme values of common standard distributions follow such a distribution and the extreme value index depends on this original law. This result shows that the distribution of the extreme values is almost independent of the distribution of the data. We can say that it is similar to the central limit theorem for the extreme values instead that for the mean value.

Let $F$ be the cumulative distribution function of $X$, i.e., $F(x)=P(X\leq x)$. The function $\Bar{F}(x)=P(X>x)$ represents the tail of the distribution of $X$. Intuitively, we can easily imagine that for most distributions the probabilities decrease when events are extreme, i.e., $\Bar{F}(x)\rightarrow0$ when x increases.
$G_{\gamma}$ tries to fit the few possible shapes for this tail. Table \ref{Tab:ExtremeDist} presents the three possible shapes of the tail and the link with the extreme value index $\gamma$. It gives also an example of a standard distribution which follows each tail behavior. Note that the parameter $\tau$ represents the bound of the initial distribution, so it could be finite or infinite.
\begin{table}[t]
\centering
\scalebox{1.0}{
\begin{tabular}{c|c|c}\hline
Tail Behavior ($x\rightarrow\tau$)&Domain&Example \\\hline
Heavy tail, $P(X>x)\simeq x^{-\frac{1}{\gamma}}$&$\gamma>0$&Frechet \\
Exponential tail, $P(X>x)\simeq e^{-x}$&$\gamma=0$&Gamma \\
Bounded, $P(X>x) \stackrel{x\geq\tau}{=}0$&$\gamma<0$&Uniform \\
\hline
\end{tabular}
}
\caption{\footnotesize Relation between $F$ and $\gamma$.}\label{Tab:ExtremeDist}
\end{table}
\begin{figure}[t]
\graphicspath{{images/}}
\centering
{\includegraphics[width=0.70\linewidth]{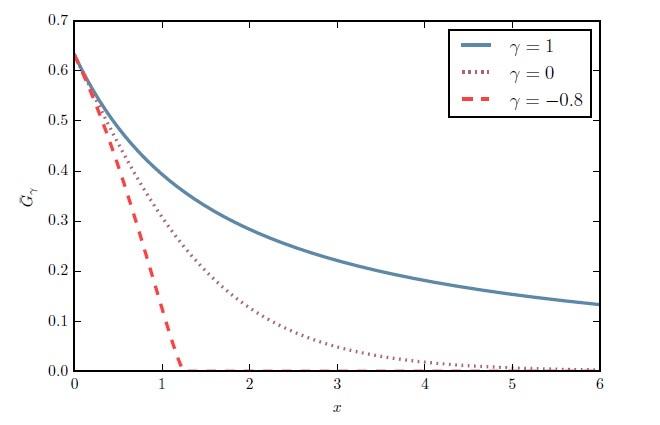}}
\caption{Tail distribution ($\Bar{G}_{\gamma}$) according to $\gamma$. This figure is in~\cite{Siffer2017}.}\label{fig:ExtremeTails}
\end{figure}
Peaks-Over-Threshold (POT) is one of the main approaches for EVT. Mathematically, for a generic random variable $X$ and a given probability $q$ we note with $k$ its quantile at level $1-q$, i.e., $k$ is the smallest value such that $\Pr(X>k)<q$. POT allows us to compute $k$ regardless of knowing the distribution of $X$. In particular, POT relies on the Pickands-Balkema-de Haan theorem~\citep{balkema1974,pickands1975}, also known as the second theorem of extreme value theory.
\begin{theorem}\label{SecondEVT}
(Pickands-Balkema-de Haan). The extreme of the cumulative distribution function $F$ converges in distribution to $G_{\gamma}$ if and only if a function $\delta$ exists, for all $x\in\mathbb{R}$ s.t. $1+\gamma x>0$:
$$\frac{\Bar{F}(l+\delta(l)x)}{F(l)}\xrightarrow[l\rightarrow\tau]{}(1+\gamma x)^{-\frac{1}{\gamma}}.$$
\end{theorem}
This theorem is also known as the second theorem of extreme value theory with respect to the initial result of Fisher, Tippett, and Gnedenko in \eqref{FirstEVT}. Shortly, for a generic threshold $l$ and a value $a$, the theorem shows that
\begin{equation}\label{SecondEVT2}
\Bar{F}(x)=P(X-l>a\ |\ X>l)\sim\left(1+\frac{\gamma x}{\delta(l)}\right)^{-\frac{1}{\gamma}}.
\end{equation}
Equation \eqref{SecondEVT2} shows that observations over a threshold $l$, written $X-l$, are likely to follow a generalized Pareto distribution (GPD) with parameters $\gamma,$ and $\delta$. In operational terms, the POT approach consists of fitting a GPD to the excesses $X-l$. To this end, we need the estimates of $\gamma$ and $\delta$ (namely, $\widehat{\gamma}$, and $\widehat{\delta}$). In our experiments, we estimated those quantities by using the maximum likelihood estimation method. After we get $\widehat{\gamma}$ and $\widehat{\delta}$, the quantile can be computed through:
\begin{equation}\label{Zt_POT}
k= l+\frac{\widehat{\delta}}{\widehat{\gamma}}\left(\left(\frac{qT}{T_l}\right)^{-\widehat{\gamma}}-1\right),
\end{equation}
where $l$ is a \enquote{high} threshold, $q$ the desired probability, $T$ the total number of observations, $T_l$ the number of peaks i.e the number of $x_t$ s.t. $x_t>l$. We set the values of $l$ and $q$ in Equation \eqref{Zt_POT} equal to the 99th percentile and 0.001, respectively. Note that in the caption of Figure \ref{tab:TabResu} of the paper we reported that some methods failed in calculating the parameters of the required generalized Pareto distribution. In particular, they obtained a non-singular Hessian matrix concerning the parameters $\widehat{\gamma}$ and $\widehat{\delta}$.

\subsection{Random Forest}\label{RandomForest}
The random forest methodology was initially proposed by \cite{Breimanrf01} as a solution to reduce the variance of regression trees~\citep{cart84} and is based on bootstrap aggregation (Bagging) of randomly constructed regression trees. This method builds several decision trees on bootstrapped training samples considering each time a random sample of $q$ predictors, is chosen as split candidates from the full set of $n$ predictors. This reduces the variance averaging many uncorrelated quantities. \\
In other terms, the random forest is a sequential procedure that splits the data in a training and validating set, called in this framework in-bag and out-of-bag samples. Then the training set for the current tree is sampled with replacement, usually, about one-third of the units are left out. The random forest method is summarized in Algorithm \ref{RFalgorithm}. Note that we set $B=100$, $T_{min}=2$ and $q=\sqrt{n}$.
The random forest algorithm allows us to obtain an overall summary of the importance of each predictor using the Gini index. In particular, let $p_{l,k}$ be the proportion of training observations in the $l$-th node that are from the $k$-th class. The Gini index is defined by
$$G=\sum_{k=1}^Kp_{l,k}(1-p_{l,k}).$$
\begin{algorithm}[H]
\begin{algorithmic}[1]
    \For{$b\gets 1, B$}
      \State (a) Draw a bootstrap sample $\mathbb{D}^*$ of size $T$ from the training data.
      \State (b) Grow a random forest tree $J_b$ to the bootstrapped data, by recursively repeating the following steps for each terminal node of
  the tree, until the minimum node size $T_{\mathrm{min}}$ is reached.
      \State \indent \makebox[2em][l]{(I)  } Select $q$ variables at random from the $n$ variables.
      \State \indent \makebox[2em][l]{(II) } Pick the best variable/split-point among the $q$.
      \State \indent \makebox[2em][l]{(III)} Split the node into two daughter nodes.
    \EndFor

    \State Output the ensemble of trees $\{J_b\}_1^B$

    \State Predict at new point $x$:
    \State Let $\hat{C}_b(x)$ be the class prediction of the $b$th random forest tree. Then $\hat{C}_r f^B(x) = \text{majority vote } \{ \hat{C}_b(x)_1^B \}$.  
  \end{algorithmic}
  \caption{Random forest for Classification}\label{RFalgorithm}
\end{algorithm}

\subsection{Relative Contribution to Deviance Explained (RCDE)}\label{RCDE}
We use the deviance of logistic regression to quantify the importance of a variable. In particular, for a generic predictor $X$, we quantify its importance by means of the Relative Contribution to Deviance Explained (RCDE)~\citep{campos2016}, that is

$$RCDE_x = \frac{(D_{null}-D_{full})-(D_{null}-D_{full\backslash X})}{(D_{null}-D_{full})}$$

where $D_{null}$ is the null deviance, $D_{full}$ is the residual deviance of the full model (including all predictors) and $D_{full\backslash X}$ is the residual deviance of the model obtained by removing $X$. The RCDE thus quantifies the percentage of the total logistic deviance attributable $X$.

\section{Precision and RIC (Training vs Test)}\label{PrecRIC_TravsTest}
\begin{figure}[H]
\graphicspath{{images/}}
\centering
{\includegraphics[width=0.8\linewidth]{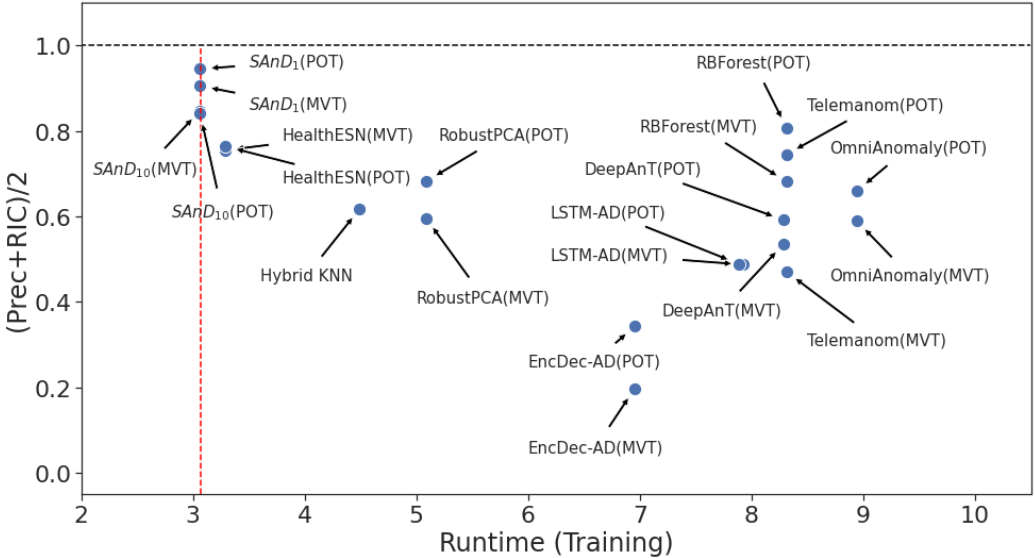}}
\caption{\footnotesize Performance for long-lived anomalies (y-axis) vs runtime (x-axis) concerning the training set, for both  MVT and POT (Hybrid KNN uses the threshold $\Pr>0.8$). 
Runtimes are averages of the logarithm of seconds needed.}\label{fig:RunTimPrecRIC_Tra}
\end{figure}

\begin{figure}[H]
\graphicspath{{images/}}
\centering
{\includegraphics[width=0.8\linewidth]{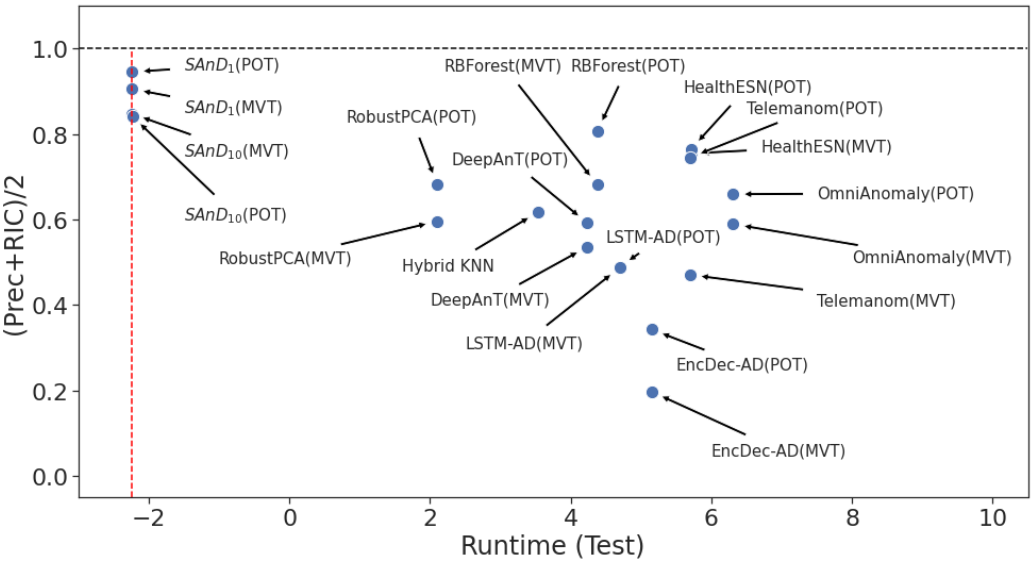}}
\caption{\footnotesize Performance for long-lived anomalies (y-axis) vs runtime (x-axis) concerning the test set, for both  MVT and POT (Hybrid KNN uses the threshold $\Pr>0.8$). 
Runtimes are averages of the logarithm of seconds needed.}\label{fig:RunTimPrecRIC_Test}
\end{figure}

Figures \ref{fig:RunTimPrecRIC_Tra} and \ref{fig:RunTimPrecRIC_Test} report the performance in detecting long-lived anomalies divided for training and test sets, respectively. Results confirm less runtime required by $\MD$.


\section{KNN}\label{KNN_others}
Table \ref{tab:TabResuHybKNN} reports the performance metrics for Hybrid KNN when the threshold is set at 0.9 and 0.99. The results show that increasing probability does not increase performance. This is because Hybrid KNN gives a high probability to both true and false positives.
\begin{table}[H]
\centering
\scalebox{.9}{
\begin{tabular}{crcrccc}
\toprule
\emph{Method}  &  \emph{MCC}  &  \emph{F1}  &  \emph{Prec}  &  \emph{RIC}  &  \emph{Rec}   &  \emph{\%No anomalies} \\ 
\midrule
\midrule
\multicolumn{7}{c}{\emph{Hybrid KNN}}\\
\midrule
\midrule
 \emph{$\Pr>0.9$}   &0.169(0.28)  &0.306(0.40)  &0.320(0.18)  &0.896(0.20)   &0.708(0.20)   &0.0\%   \\
 \emph{$\Pr>0.99$}  &0.274(0.31)  &0.336(0.43)  &0.384(0.22)  &0.855(0.22)   &0.625(0.21)   &0.0\%   \\
\bottomrule
\end{tabular}
}

\caption{\footnotesize Performance metrics of Hybrid KNN with probability threshold at 0.9 and 0.99.}\label{tab:TabResuHybKNN}
\end{table}

\section{RIC}\label{RIC}
Assume that $J$ long-lived anomalies occur in $\mathbf{X}_B$. Let $TP_j$ be the number of true positives in the confusion matrix relative to long-lived anomaly $j$, $j=1\dots,J$. We consider the function $f(TP_j)=1$ if $TP_j>0$ and $f(TP_j)=0$ otherwise. Let a long-lived anomaly be indicated as a cluster, we define the ratio of identified clusters ($RIC$), as $RIC=\frac{\sum_j f(TP_j)}{J}$.

\section{Focusing on long-lived anomalies }\label{TestPLOT}
\begin{figure}[H]
	\graphicspath{{images/}}
	\centering
	\subfloat[]{\includegraphics[height=4.4cm]{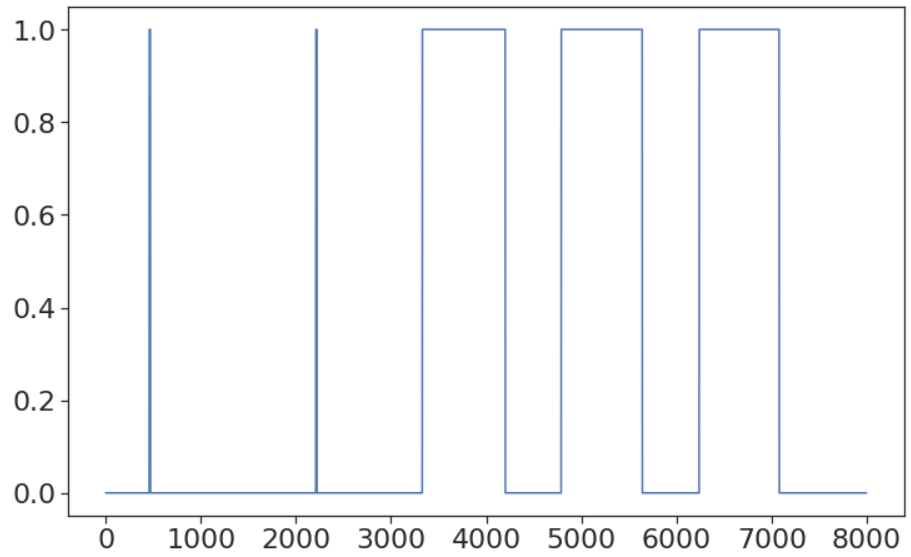}}\hfill
	\subfloat[]{\includegraphics[height=4.4cm]{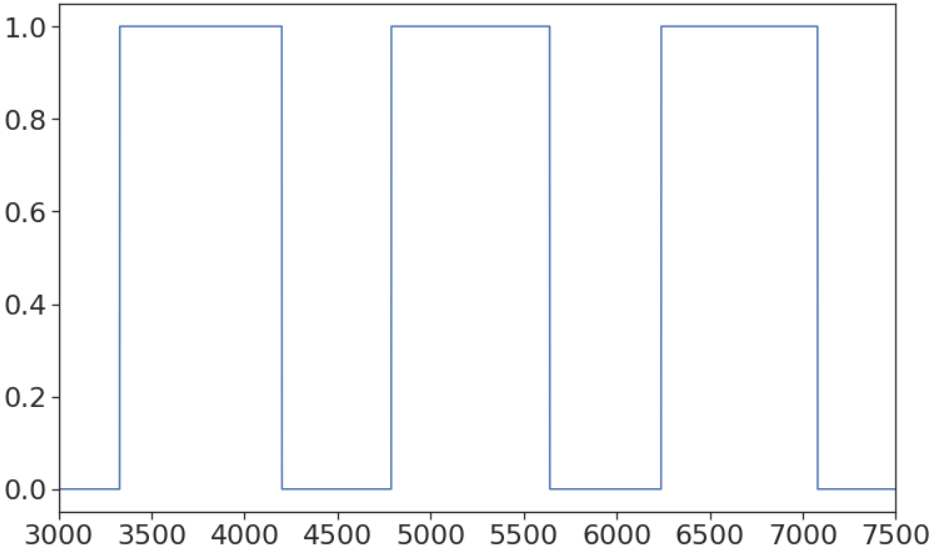}}
\caption{\footnotesize Anomalies on the test set of the SMD 2-2 dataset. Plot (a) reports all the anomalies. Plot (b) reports the long-lived ones.}\label{fig:figTestPLOT_SMD22}
\end{figure}

\begin{figure}[H]
	\graphicspath{{images/}}
	\centering
	\subfloat[]{\includegraphics[height=4.4cm]{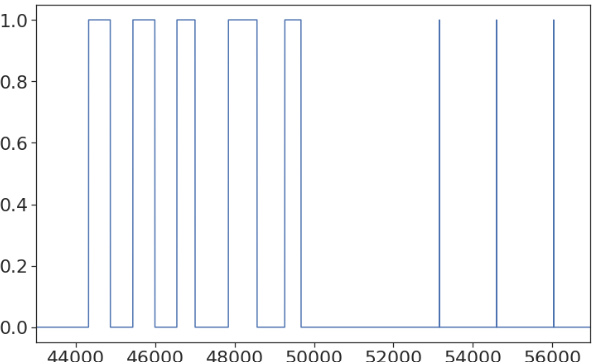}}\hfill
	\subfloat[]{\includegraphics[height=4.4cm]{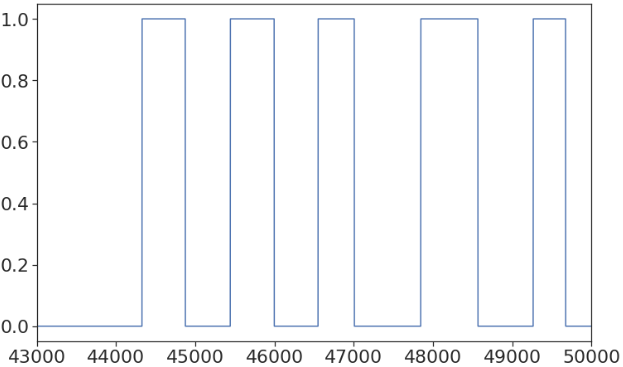}}
\caption{\footnotesize Anomalies on the test set of the SMD 1-1 dataset. Plot (a) reports all the anomalies. Plot (b) reports the long-lived ones.}\label{fig:figTestPLOT_SMD11}
\end{figure}

\begin{figure}[H]
	\graphicspath{{images/}}
	\centering
	\subfloat[]{\includegraphics[width=5cm]{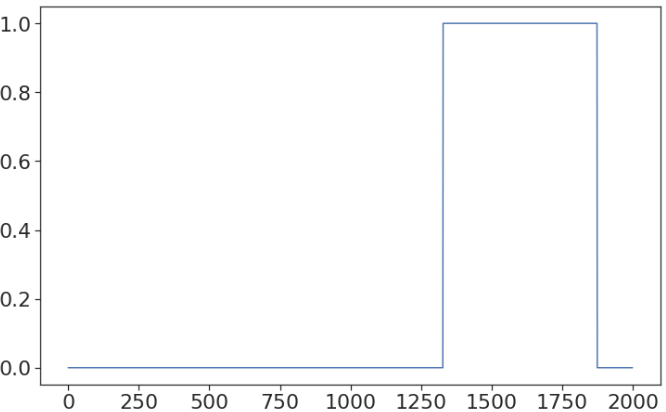}}\hfil
	\subfloat[]{\includegraphics[width=5cm]{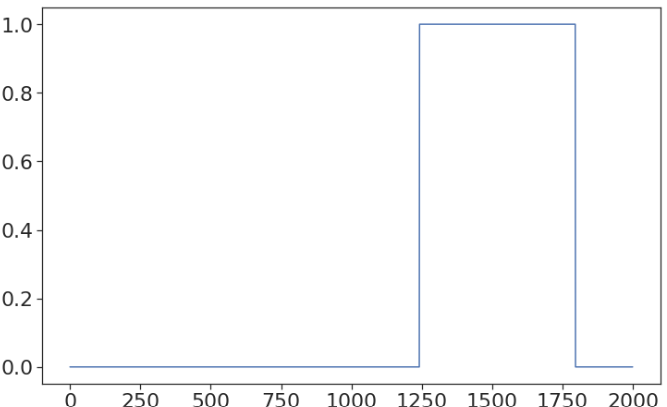}}\hfil 
	\subfloat[]{\includegraphics[width=5cm]{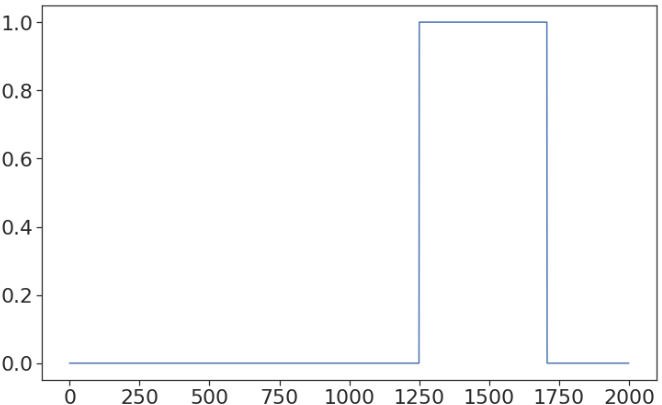}}\hfill
	
	\subfloat[]{\includegraphics[width=5cm]{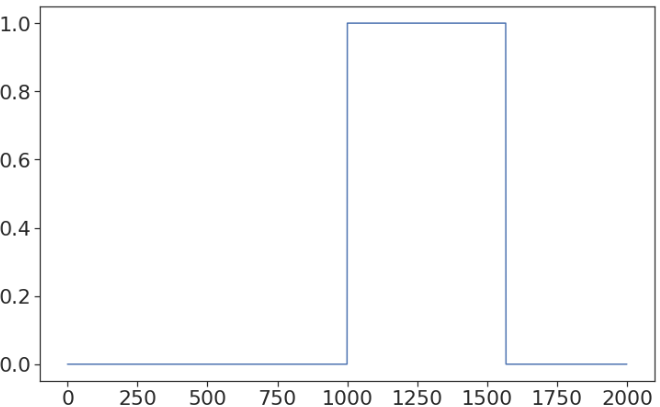}}\hfil 
	\subfloat[]{\includegraphics[width=5cm]{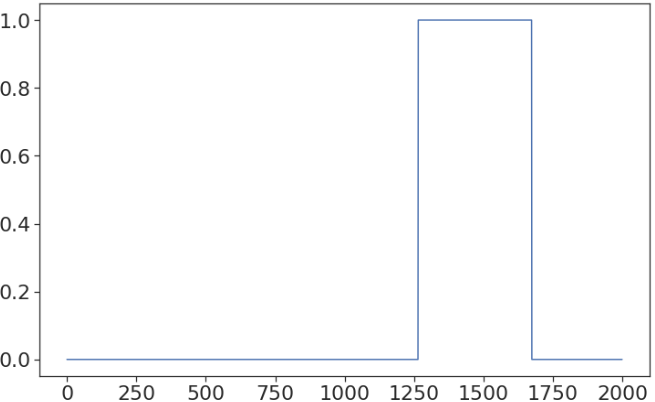}}\hfil
	\caption{Panels
		(a)-(e) show the 5 datasets of size \np{2000} (\np{1000} observations from the test set and \np{1000} additional observations from the training set) concerning the long-lived anomlies of the 1-1 SMD dataset.}\label{fig:longlivedSMD11}
\end{figure}

Here is an example of how we selected long-lived anomalies. Figure~\ref{fig:figTestPLOT_SMD22} shows the anomalies in the test set for the SMD 2-2 dataset (the second dataset that we considered from the SMD collection). Plot (a) shows the total anomalies, 2 short-lived and 3 long-lived. In our analysis, we focused on the anomalies shown in plot (b). This selection was repeated for all the 8 datasets considered. Figure~\ref{fig:figTestPLOT_SMD11} shows the anomalies in the test set for the SMD 1-1 dataset. Plot (a) shows the 8 anomalies, while plot (b) shows the 5 long-lived of interest in Section~\ref{sec:rootCause}. Figure~\ref{fig:longlivedSMD11} reports in each plot the data used to obtain the results in Table~\ref{Tab:VarSel}.

\section{Feature Importance}\label{FeatureImportance}
\begin{figure}[H]
	\graphicspath{{images/}}
	\centering
	\subfloat[Anomaly 1]{\includegraphics[width=5cm]{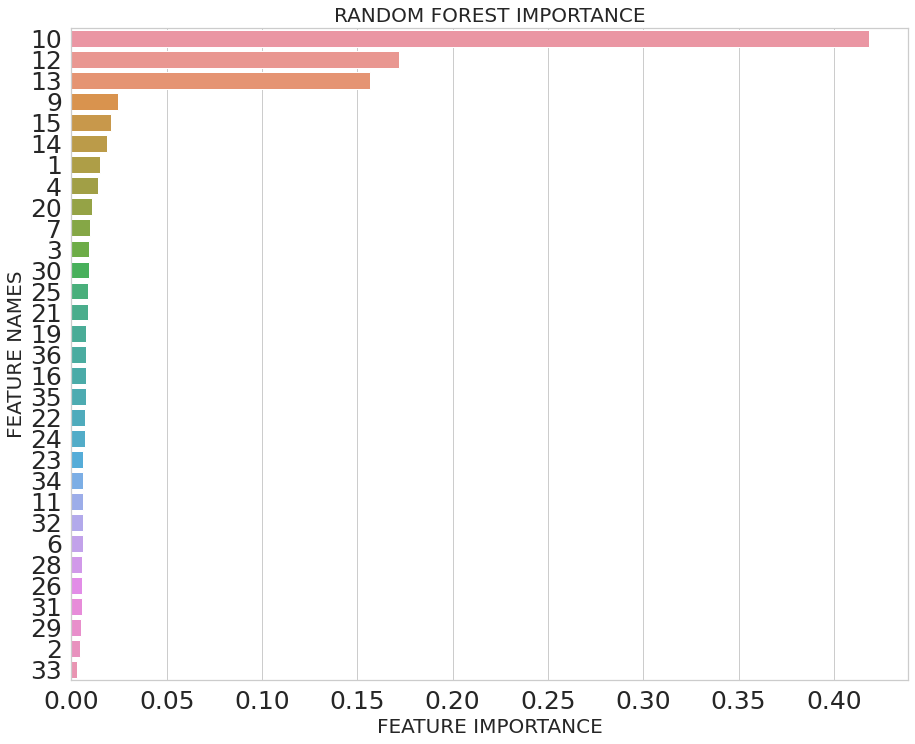}}\hfil
	\subfloat[Anomaly 2]{\includegraphics[width=5cm]{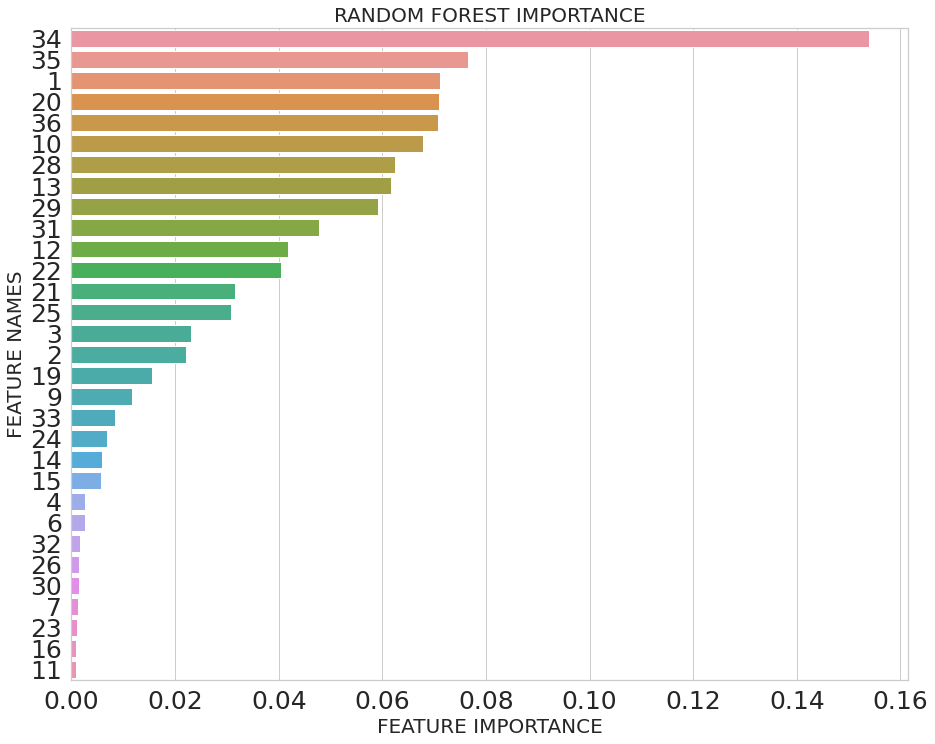}}\hfil   
 	\subfloat[Anomaly 3]{\includegraphics[width=5cm]{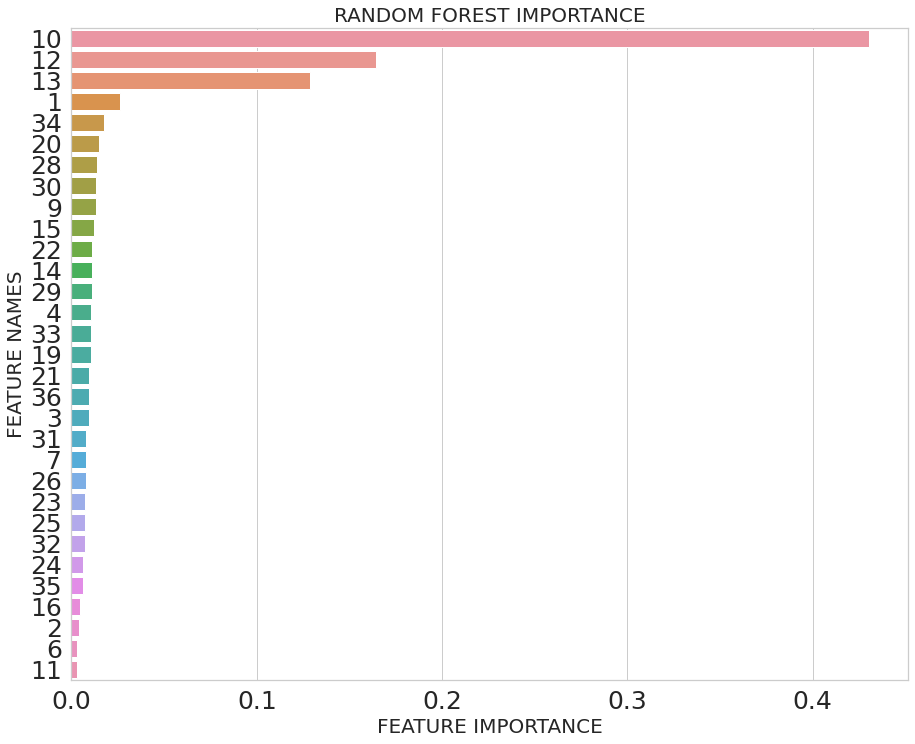}}\hfil   
	
	\subfloat[Anomaly 4]{\includegraphics[width=5cm]{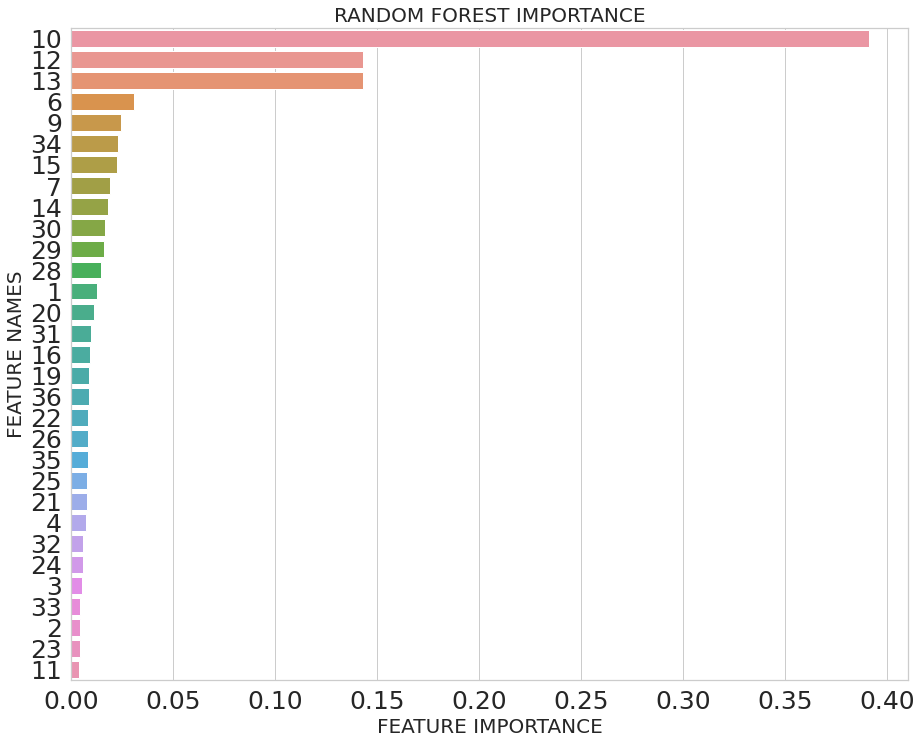}}\hfil 
	\subfloat[Anomaly 5]{\includegraphics[width=5cm]{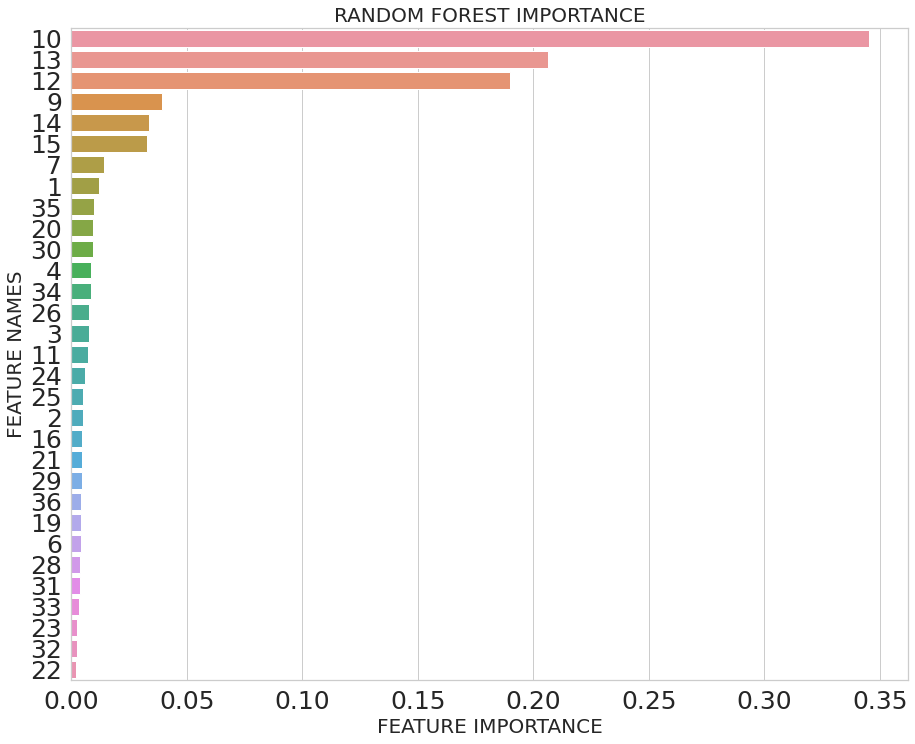}}\hfil
	\caption{\footnotesize Variables importance as computed by our methodology for the 5 long-lived anomalies for $h=1$ (Anomaly 1 - Anomaly 5). The $x$-axis shows the value of the Gini index. The $y$-axis reports the label of the variables in increasing order of the Gini index.}\label{fig:VariableImport_h1}
\end{figure}

\begin{figure}[H]
	\graphicspath{{images/}}
	\centering
	\subfloat[Anomaly 1]{\includegraphics[width=5cm]{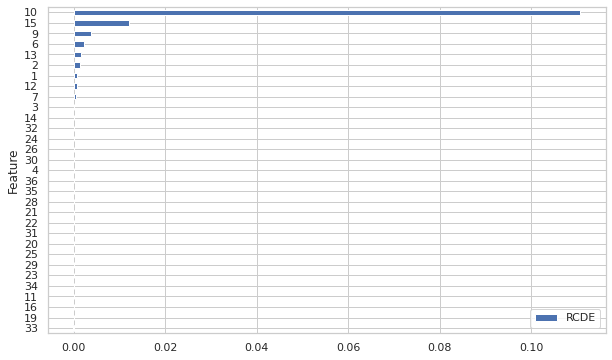}}\hfil
	\subfloat[Anomaly 2]{\includegraphics[width=5cm]{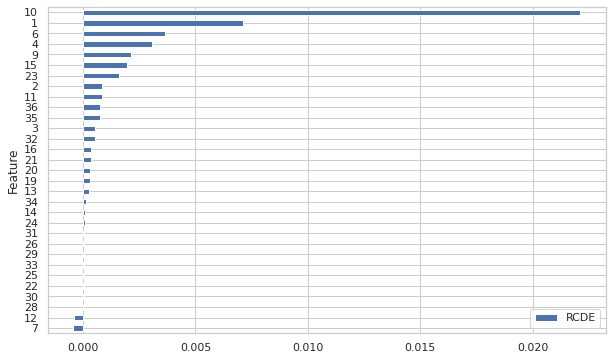}}\hfil   
 	\subfloat[Anomaly 3]{\includegraphics[width=5cm]{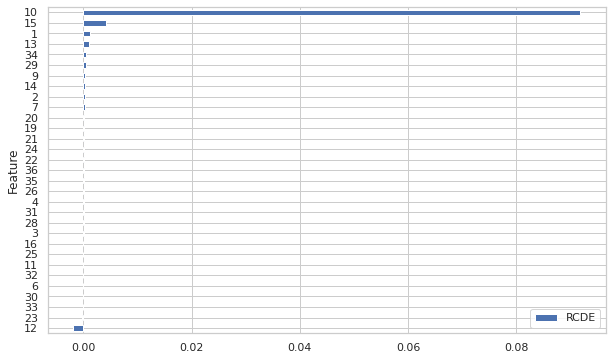}}\hfil   
	
	\subfloat[Anomaly 4]{\includegraphics[width=5cm]{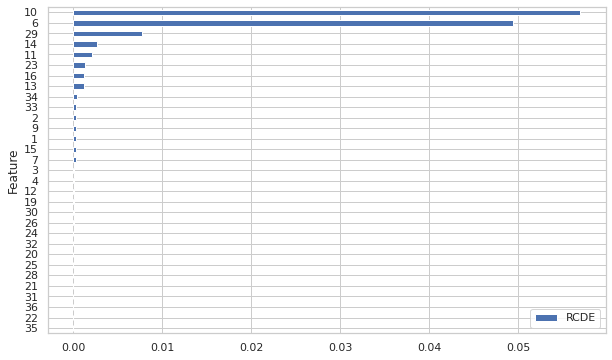}}\hfil 
	\subfloat[Anomaly 5]{\includegraphics[width=5cm]{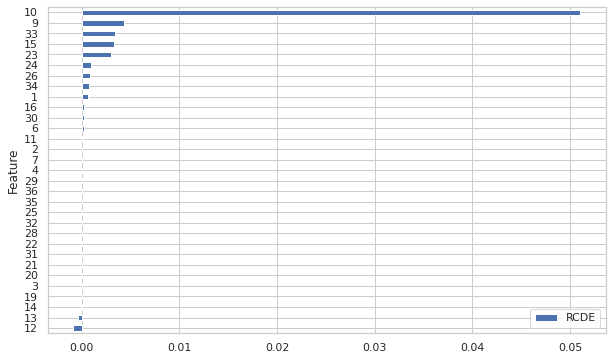}}\hfil
	\caption{\footnotesize Variables importance as computed by our methodology for the 5 long-lived anomalies for $h=1$ (Anomaly 1 - Anomaly 5). The $x$-axis shows the value of the RCDE. The $y$-axis reports the label of the variables in increasing order of the RCDE.}\label{fig:VariableImportRCDE_h1}
\end{figure}

\begin{figure}[H]
	\graphicspath{{images/}}
	\centering
	\subfloat[Anomaly 1]{\includegraphics[width=5cm]{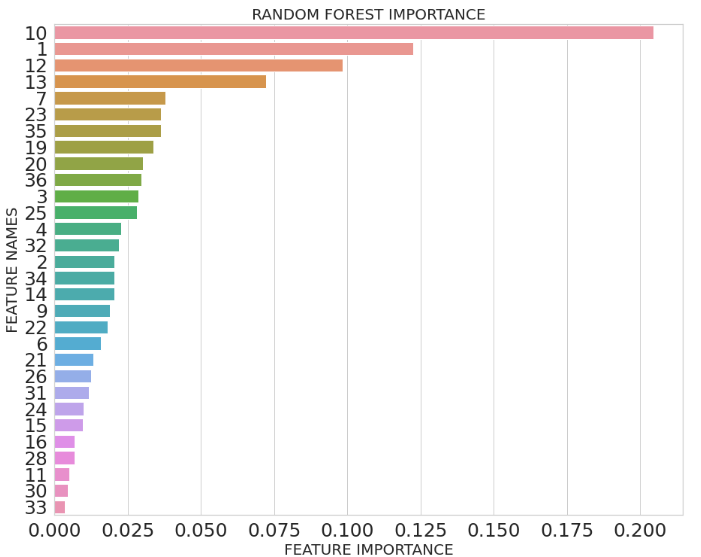}}\hfil
	\subfloat[Anomaly 2]{\includegraphics[width=5cm]{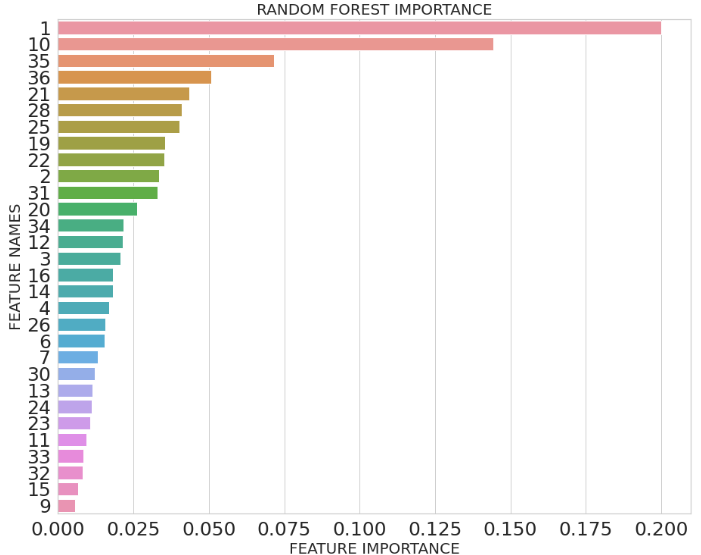}}\hfil   
 	\subfloat[Anomaly 3]{\includegraphics[width=5cm]{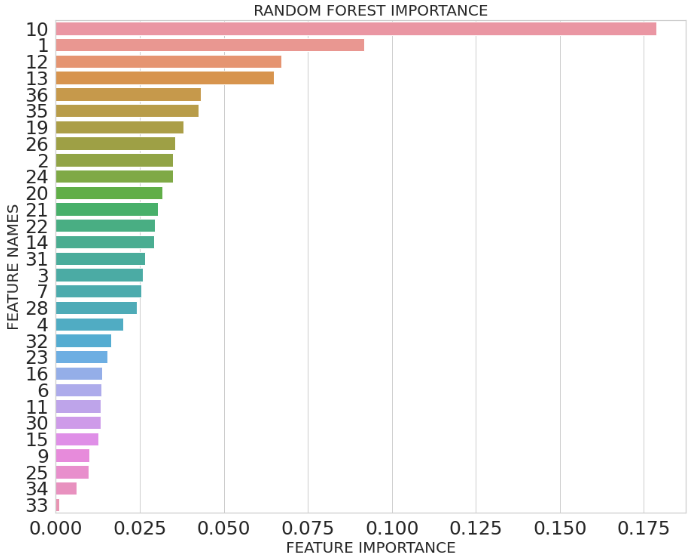}}\hfil   
	
	\subfloat[Anomaly 4]{\includegraphics[width=5cm]{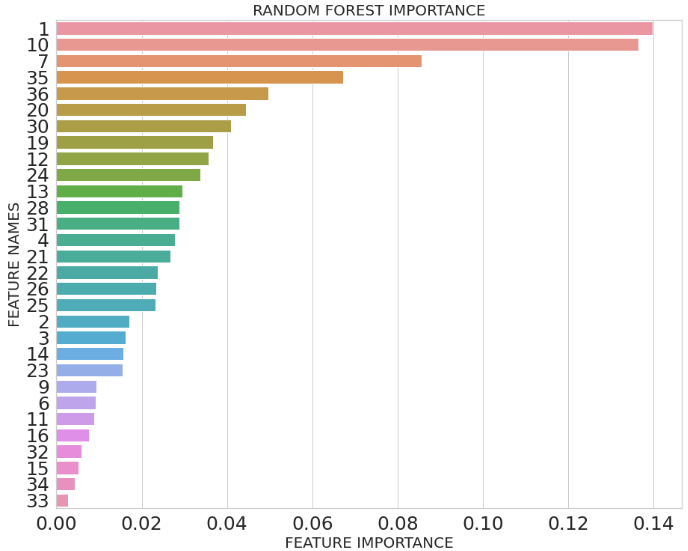}}\hfil 
	\subfloat[Anomaly 5]{\includegraphics[width=5cm]{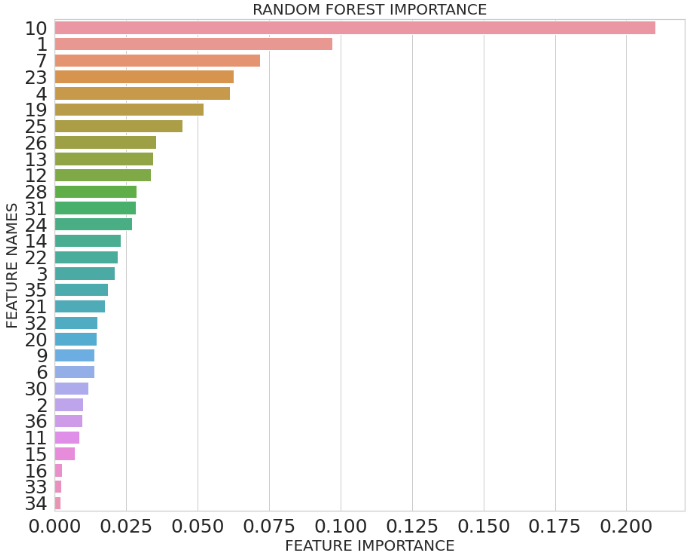}}\hfil
	\caption{\footnotesize Variables importance as computed by our methodology for the 5 long-lived anomalies for $h=10$ (Anomaly 1 - Anomaly 5). The $x$-axis shows the value of the Gini index. The $y$-axis reports the label of the variables in increasing order of the Gini index.}\label{fig:VariableImport_h10}
\end{figure}

\begin{figure}[H]
	\graphicspath{{images/}}
	\centering
	\subfloat[Anomaly 1]{\includegraphics[width=5cm]{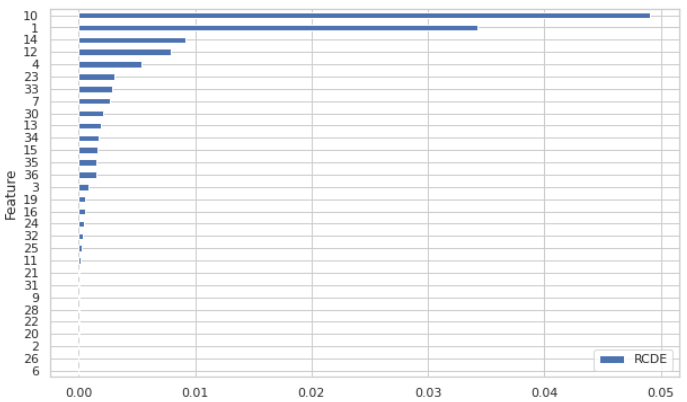}}\hfil
	\subfloat[Anomaly 2]{\includegraphics[width=5cm]{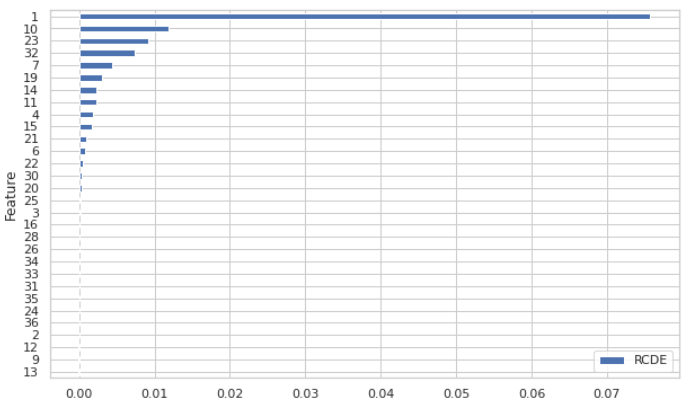}}\hfil   
 	\subfloat[Anomaly 3]{\includegraphics[width=5cm]{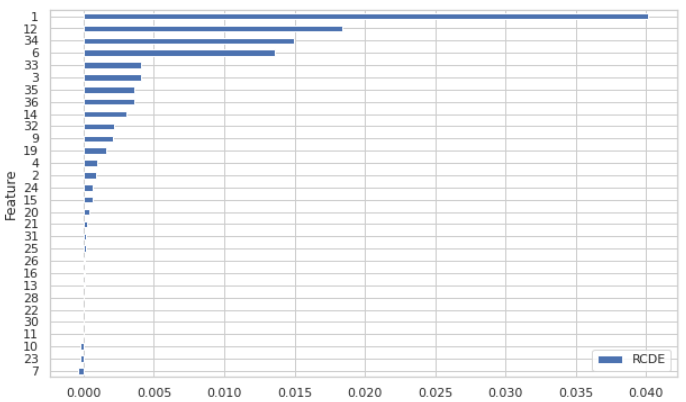}}\hfil   
	
	\subfloat[Anomaly 4]{\includegraphics[width=5cm]{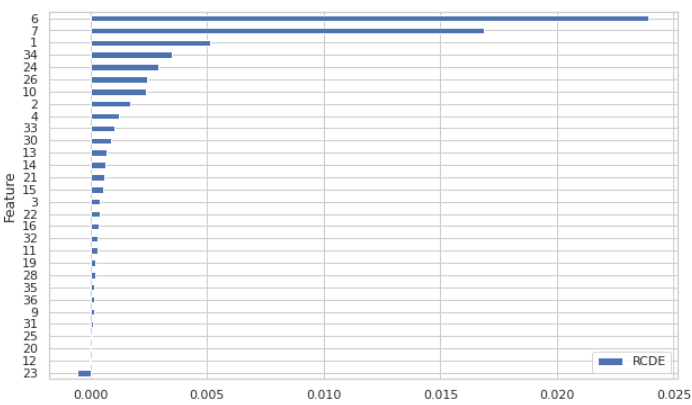}}\hfil 
	\subfloat[Anomaly 5]{\includegraphics[width=5cm]{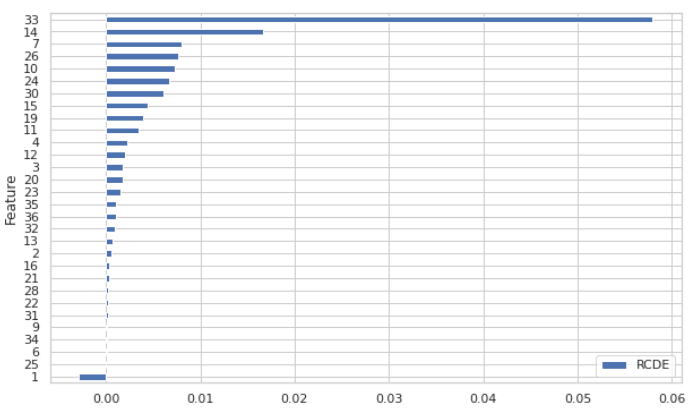}}\hfil
	\caption{\footnotesize Variables importance as computed by our methodology for the 5 long-lived anomalies for $h=10$ (Anomaly 1 - Anomaly 5). The $x$-axis shows the value of the RCDE. The $y$-axis reports the label of the variables in increasing order of the RCDE.}\label{fig:VariableImportRCDE_h10}
\end{figure}

Figure 4 reports the ranking of the Gini index provided by the random forest in the case of $h=1$ (the higher is the value, the more important is the variable). It might be interesting to use this value to automatically decide how many variables $n$ to report to the user. For example, Figure~\ref{fig:VariableImport_h1} (a),(c),(d),(e) show that the variables 10, 12 and 13 are the most relevant for the corresponding anomalies. Similarly, Figure~\ref{fig:VariableImportRCDE_h1} reports the RCDE for $h=1$. Figures~\ref{fig:VariableImport_h10} and~\ref{fig:VariableImportRCDE_h10} report the Gini index and the RCDE rankings, respectively, in the case of $h=10$.

\section{Analysis of the Short-Lived Anomaly for POT and $h=10$}\label{ShortPOTh10}
\begin{figure}[H]
	\graphicspath{{images/}}
	\centering
	{\includegraphics[width=0.6\linewidth]{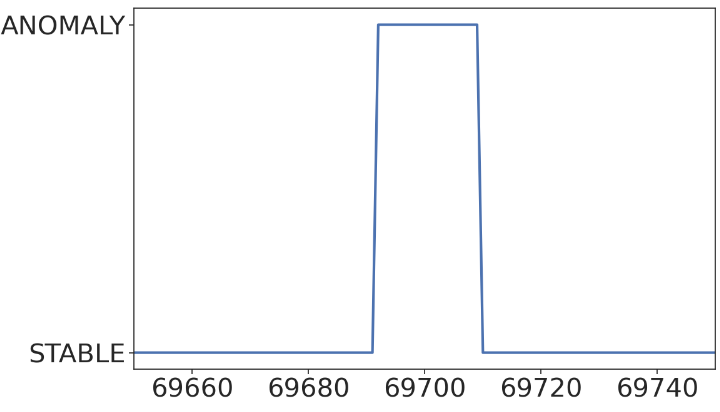}}\hfill
 
	\caption{\footnotesize The short-lived anomaly detected by $\MD_{10}$ and reported in Figure~\ref{fig:ShortLivPOTh10} (c).}\label{fig:ShortLivPOTh10}
\end{figure}

Figure \ref{fig:ShortLivPOTh10} reports the short-lived anomaly obtained by running $\MD_{10}$ thresholded with POT. It lasts 20 seconds, so it is not due to sensor noises that last only one second. We run a random forest on the 100 observations reported in Figure~\ref{fig:ShortLivPOTh10}, to maintain a ratio between the anomaly and the anomaly-free observations $\approx0.25$, as in the case of the long-lived anomaly in Figure \ref{fig:Celli2} (d). The 5 top variables are 
\textit{ReelPower},
\textit{DR1PowerConsumption},
\textit{ReelSpeed},
\textit{MCC4PowerConsumption}, and
\textit{YankePressure}. 
All these variables are still related to overheating of the system.

\end{appendices}

\end{document}